\pdfoutput=1

\documentclass[11pt]{article}

\usepackage{aclpgnm}

\usepackage{comment}

\usepackage[OT1]{fontenc}\usepackage{times}
\usepackage{latexsym}
\usepackage{graphicx}
\usepackage{amsmath}
\usepackage{booktabs, multirow} %
\usepackage{soul}%
\usepackage{hieroglf}

\usepackage{nth}

\usepackage{CJKutf8}
\newcommand{\inlinejp}[1]{\begin{CJK}{UTF8}{min}{#1}\end{CJK}}
\newcommand{\inlinezh}[1]{\begin{CJK}{UTF8}{gbsn}{#1}\end{CJK}}

\usepackage{microtype}

\usepackage{inconsolata}

\title{Lost in Translation? Translation Errors and Challenges for Fair Assessment of Text-to-Image Models on Multilingual Concepts}

\author{Michael Saxon$^{\textpmhg{eF}}$\quad Yiran Luo$^{\textpmhg{eR}}$\quad Sharon Levy$^{\textpmhg{\Hibl}}$ \\ \textbf{Chitta Baral$^{\textpmhg{R}}$\quad Yezhou Yang$^{\textpmhg{R}}$\quad William Yang Wang$^{\textpmhg{F}}$}\\
$^{\textpmhg{F}}$University of California, Santa Barbara \; $^{\textpmhg{R}}$Arizona State University \; $^{\textpmhg{\Hibl}}$Johns Hopkins University\\
$^{\textpmhg{e}}$\textit{Equal contribution \& corresponding}: \texttt{\href{mailto:saxon@ucsb.edu}{saxon@ucsb.edu}, \href{mailto:yluo97@asu.edu}{yluo97@asu.edu}}
}

\begin{document}
\maketitle

\begin{abstract}
Benchmarks of the multilingual capabilities of text-to-image (T2I) models compare generated images prompted in a test language to an expected image distribution over a concept set. One such benchmark, ``Conceptual Coverage Across Languages'' (CoCo-CroLa), assesses the tangible noun inventory of T2I models by prompting them to generate pictures from a concept list translated to seven languages and comparing the output image populations. Unfortunately, we find that this benchmark contains translation errors of varying severity in Spanish, Japanese, and Chinese. We provide corrections for these errors and analyze how impactful they are on the utility and validity of CoCo-CroLa as a benchmark. We reassess multiple baseline T2I models with the revisions, compare the outputs elicited under the new translations to those conditioned on the old, and show that a correction's impactfulness on the image-domain benchmark results can be predicted in the text domain with similarity scores. Our findings will guide the future development of T2I multilinguality metrics by providing analytical tools for practical translation decisions.
\end{abstract}

\section{Introduction}

With growth in the popularity of generative text-to-image (T2I) models has come interest in assessing their capabilities across many dimensions, including multilingual accessibility. 
The CoCo-CroLa \cite{saxon-wang-2023-multilingual} benchmark attempts to capture how well ``concept-level knowledge'' within a T2I model is accessible across different input languages.
It compares the output image populations of a system under test when prompted to generate images of 193 tangible concepts in 7 \textit{test languages} to the images generated from a semantically equivalent prompt in a \textit{source language}. 
It and similar benchmarks rely on correct translations for validity, lest ``possessed'' concepts be mistakenly assigned false negatives.

\begin{figure}[!t]
\includegraphics[width=\linewidth]{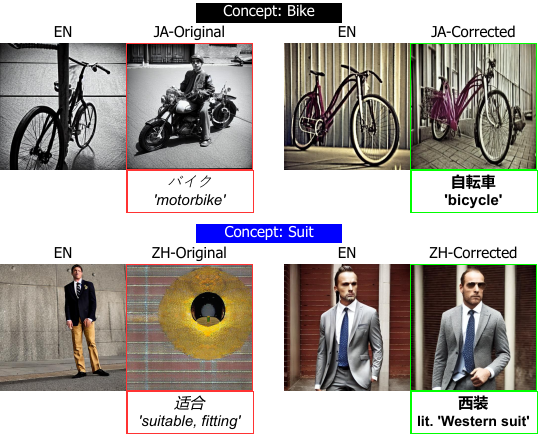}
\caption{%
The CoCo-CroLa benchmark mistranslated concepts such as \textit{bike} in JA and \textit{suit} in ZH. With correct translations (right) AltDiffusion does in fact ``possess'' them; originally (left) they were false negatives.
}
\label{fig:figure1}
\vspace{-12pt}
\end{figure}

\begin{table*}[!t]\centering
\scriptsize
\begin{tabular}{llllll}\toprule
Concept&Language &Original &Corrected &Reason for Correction \\ \midrule
Rock &Japanese &\inlinejp{ロック} &\inlinejp{岩} &\inlinejp{ロック}, \textit{rokku}, refers principally to ``rock music'' instead of stones in nature. \\
Flame &Spanish &\textit{llama} &\textit{flama} &\textit{Llama}, though a correct translation for ``flame,'' coincides with the animal in English. \\
Ground &Japanese &\inlinejp{接地} &\inlinejp{地面} &\inlinejp{接地} refers to an electrical ground rather than the surface of the earth. \\
Table &Chinese &\inlinezh{表} &\inlinezh{桌子} &\inlinezh{表} means a tabular form or a spreadsheet, not a four-legged furniture. \\
Milk &Japanese &\inlinejp{乳} &\inlinejp{牛乳} &\inlinejp{乳} may mean breast or any kind of milk. \inlinejp{牛乳} means the milk produced by cows. \\
Tent &Spanish &\textit{tienda} &\textit{...de acampar} & \textit{Tienda} alone more often means ``store,'' \textit{tienda de acampar} specifies (camping) tent. \\
Teacher &Japanese &\inlinejp{先生} &\inlinejp{教師} &\inlinejp{先生} is a common title to address an educated person, e.g., teacher, doctor, lawyer. \\
Father &Chinese &\inlinezh{爸爸} &\inlinezh{父亲} &\inlinezh{爸爸} is the colloquial addressing equivalent to `daddy'. \inlinezh{父亲} is more formal. \\
\bottomrule
\end{tabular}
\caption{Example error candidates from the CoCo-CroLa benchmark in Japanese, Chinese, and Spanish. \label{tab: translation-errors}
}
\end{table*}

We find a strict \textit{error candidate rate} of 4.7\% for Spanish (ES), 8.8\% for Chinese (ZH), and 12.9\% for Japanese (JA) in the CoCo-CroLa v1 (CCCL) concept translations through manual analysis by fluent speakers.
These error candidates are not filtered by severity. 
While some candidates are severe translation errors that drive false negatives (\autoref{fig:figure1}), 
others are marginal annotator disagreements that might not matter (\autoref{tab: translation-errors}). 
In this work, \textbf{we investigate when and why translation changes actually impact CCCL results} to improve future T2I multilinguality benchmarks. We:
\begin{enumerate}
    \item Write \textit{candidate corrections} for CCCL in ES, JA, and ZH, evaluated on four T2I models.
    \item Introduce a text-domain comparison metric $\Delta\mathrm{SEM}$ to predict correction significance. 
    \item Analyze our candidates by $\Delta\mathrm{SEM}$ and image correctness improvement and apply impactful ones to CCCL as v1.1.
    \item Report insights and considerations for future semantic T2I evaluations we uncovered.
\end{enumerate}

\section{Motivation \& Approach}

The \textbf{C}o\textbf{C}o-\textbf{C}ro\textbf{L}a benchmark (CCCL) evaluates a T2I model's ability to generate images of an inventory of tangible concepts when prompted in different languages \cite{saxon-wang-2023-multilingual}. 
Given a tangible concept $c$, written in language $\ell$ as phrase $c_\ell$, the $i$-th image produced by a multilingual T2I model $f$ on the concept $c_\ell$ can be expressed as:
\begin{equation}
I_{c_\ell,i} \sim f(c_\ell)
\label{eq:I}
\end{equation}

The images generated in language $\ell$ are considered \textit{correct} if they are faithful to their equivalent counterparts in the source language $\ell_s$. This is measured by the CCCL benchmark by a \textbf{correctness metric} for a single concept $c$ as the \textit{cross-consistency} score $X_c (f, c_\ell, c_{\ell_s})$:
\begin{equation}
X_c = \frac{1}{n^2}\sum^{n}_{i=0}\sum^{n}_{j=0}\mathrm{SIM}_{F}(I_{c_\ell,i}, I_{c_{\ell_s},j})
\label{eq:xc}
\end{equation}
where we sample $n$ images per-concept per-language (we use 9), and $\mathrm{SIM}_{F}(\cdot,\cdot)$ measures the cosine similarity in feature space by image feature extractor $F$. In practice, the default source language $\ell_s$ is English and $F$ is the CLIP visual feature extractor \citep{radford2021learning}. 

\subsection{Translation Errors in CoCo-CroLa}

CCCL requires correct translations of each concept $c$ from the source language $\ell_s$ into a set of semantically-equivalent translations in each test language $\ell$. 
\citet{saxon-wang-2023-multilingual} built CCCL v1's concept translation list using an automated approach so as to allow new languages to be easily added without experts in each new language.

They used an ensemble of commercial machine translation systems to generate candidate translations and the BabelNet knowledge graph \cite{navigli2010babelnet} to enforce word sense agreement. Unfortunately, this approach introduces translation errors (\autoref{tab: translation-errors}). 

We check the Spanish, Chinese, and Japanese translations using a group of proficient speakers, following a protocol described in Appendix \ref{sec:procedure}, who identify a set of \textit{translation error candidates} that may not sufficiently capture a concept's intended semantics in English, for various reasons.

Some of the candidate errors, such as the error for \textit{rock} in JA (\autoref{tab: translation-errors}), represent severe failures to translate a concept into its common, tangible sense---it is incoherent to test a model's ability to generate pictures of rocks by prompting it with ``rock music.'' However, other candidate errors, such as \textit{father} in ZH are still potentially acceptable translations, but deviate from the annotators' preferred level of formality or specificity.

To decide which corrections ought to be integrated in future T2I multilinguality benchmarks, 
quantifying both the significance of each translation correction is and its impact on the CCCL score for its concept is desirable.

\subsection{Quantifying Error Correction \& Impact}

Characterizing the \textit{impact} of a translation correction on model behavior is simple; we check $\Delta X_c$, the change in the CCCL score going from the original concept translation $c_\ell$ to the corrected $c'_\ell$,

\begin{equation}
    \Delta X_c(c, \ell) = X_c(f,c'_\ell,c_{\ell_s}) - X_c(f,c_\ell,c_{\ell_s})
\end{equation}

\noindent by comparing the generated population of images elicited from the corrected term $I_{c'_{\ell}}$ to the candidate translation error-conditioned images  $I_{c_{\ell}}$.

We quantify the significance of the translation correction as the \textit{improvement in semantic similarity} $\Delta\mathrm{SEM}(c_{\ell_s},c_{\ell},c'_{\ell})$ using a text feature extractor $F_t$ and cosine similarity metric $\mathrm{SIM}(\cdot,\cdot)$

\begin{equation}
    \Delta\mathrm{SEM} = \mathrm{SIM}_{Ft}(c_{\ell_s},c'_{\ell}) - \mathrm{SIM}_{Ft}(c_{\ell_s},c_{\ell})
\end{equation}

We use embeddings from the multilingual SentenceBERT \cite{reimers-2019-sentence-bert} text embedder \texttt{OpenAI CLIP-ViT-B32} model as $F_t$.

\begin{figure*}[!t]
\centering
\includegraphics[width=0.33\linewidth]{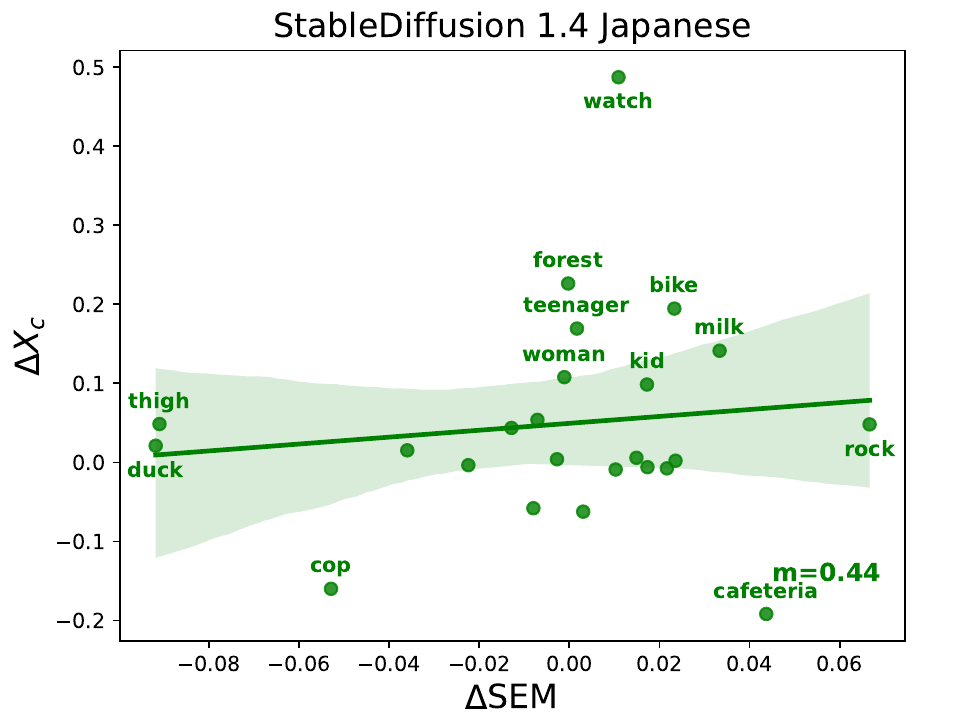}\includegraphics[width=0.33\linewidth]{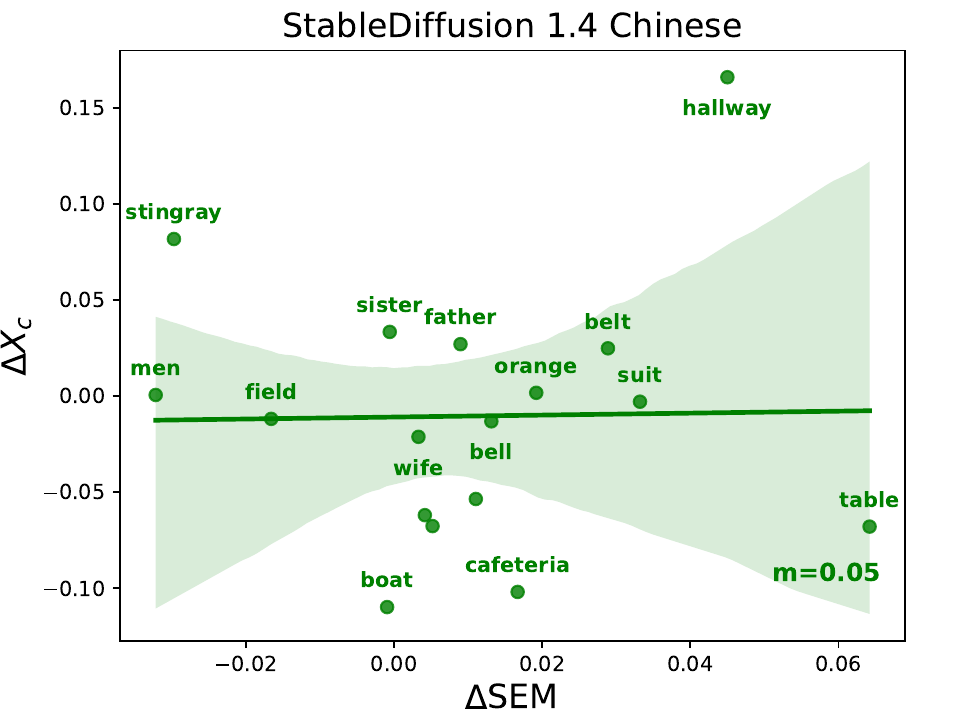}\includegraphics[width=0.33\linewidth]{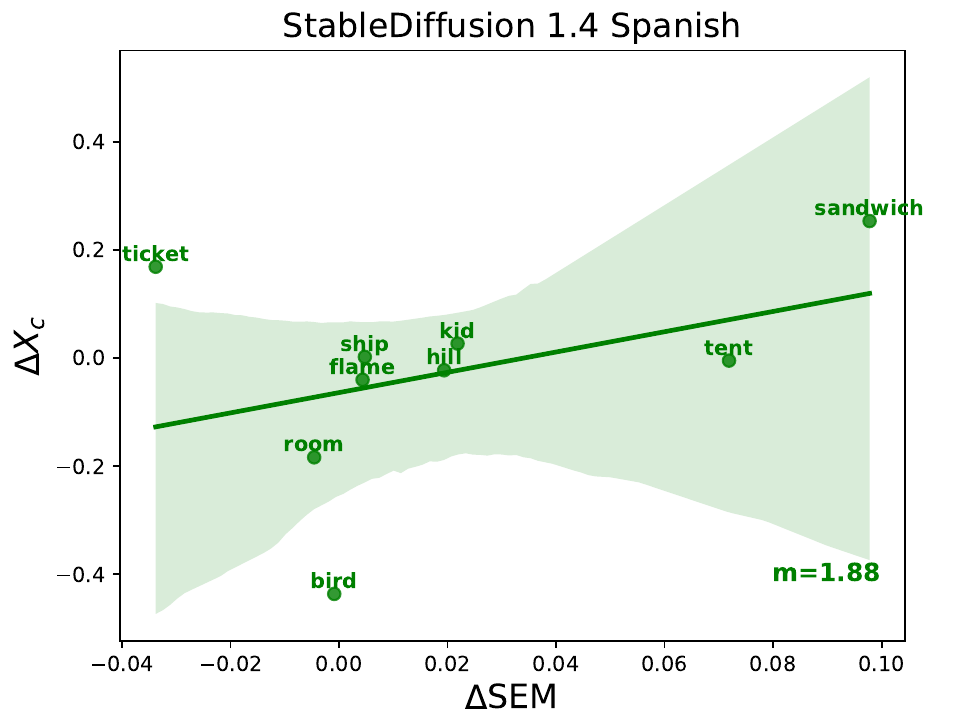} \\
\includegraphics[width=0.33\linewidth]{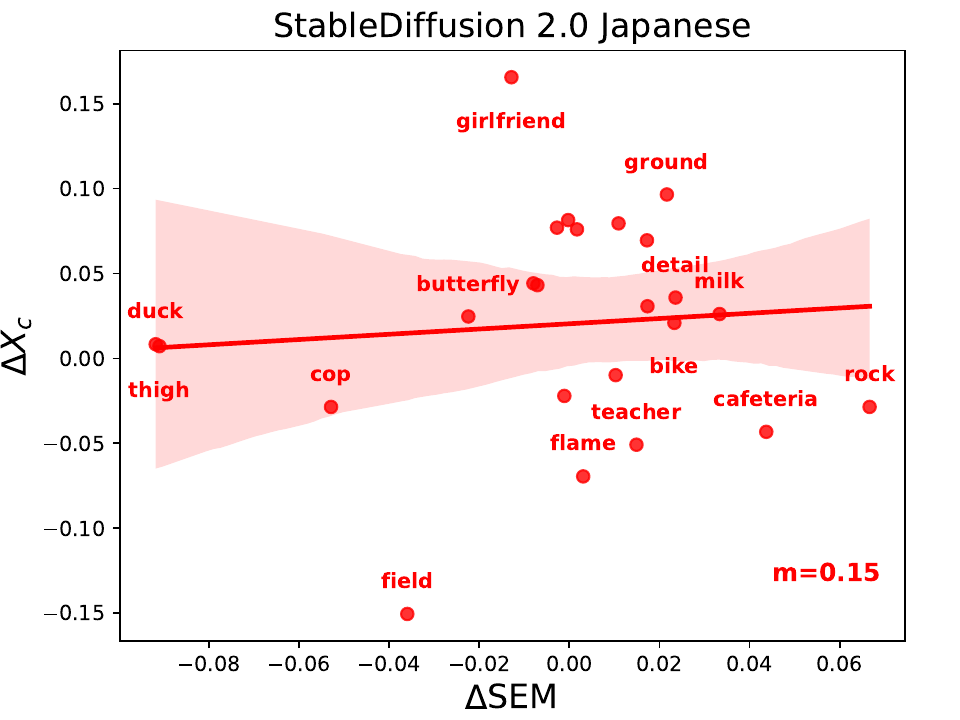}\includegraphics[width=0.33\linewidth]{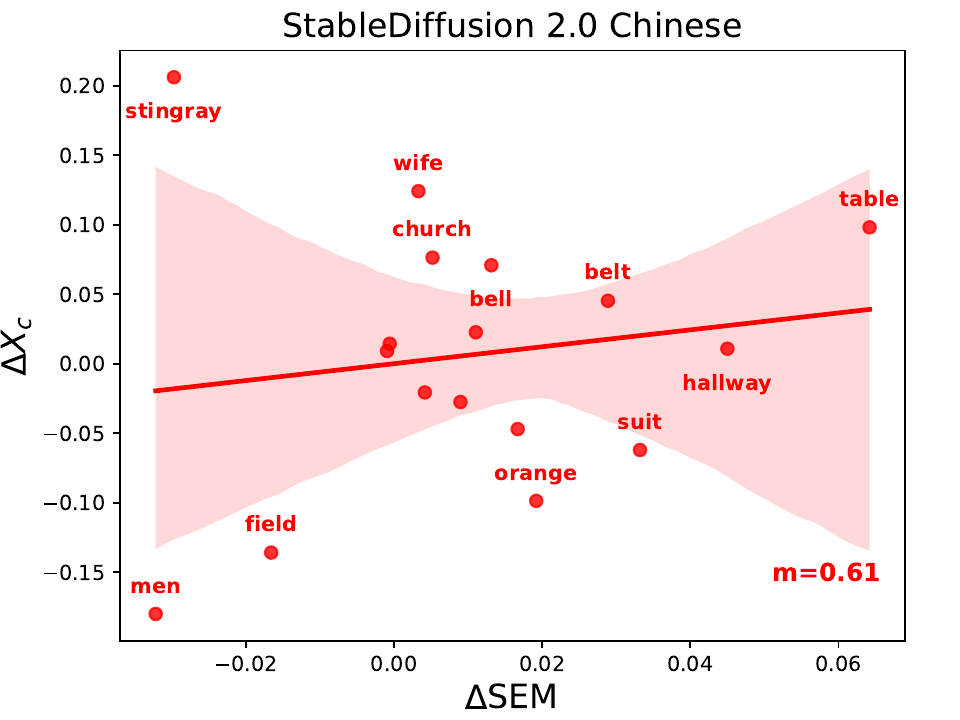}\includegraphics[width=0.33\linewidth]{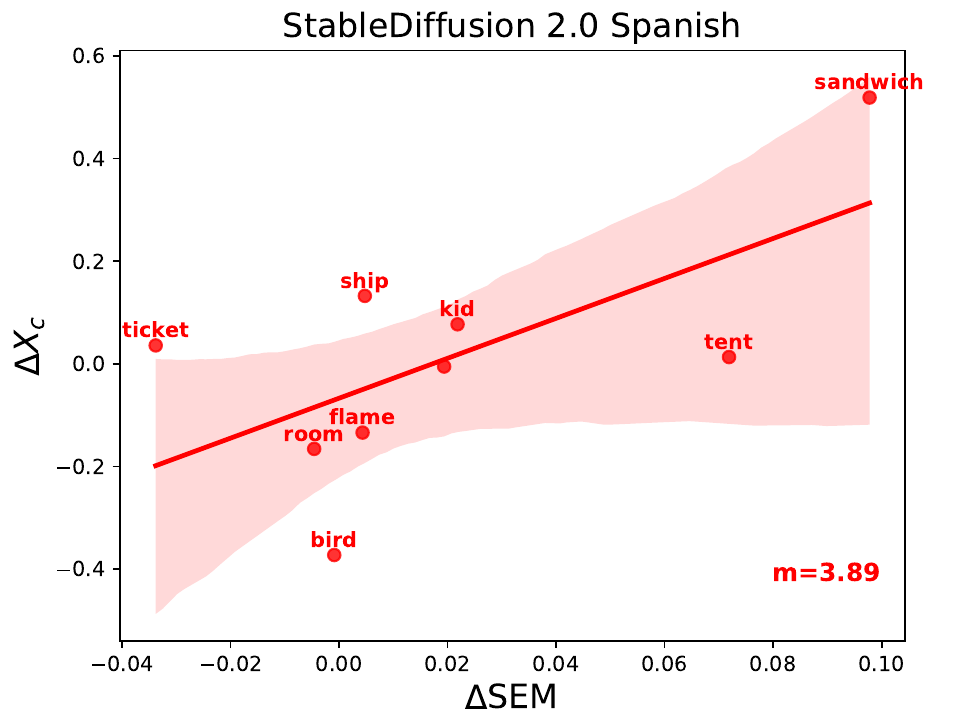} \\
\includegraphics[width=0.33\linewidth]{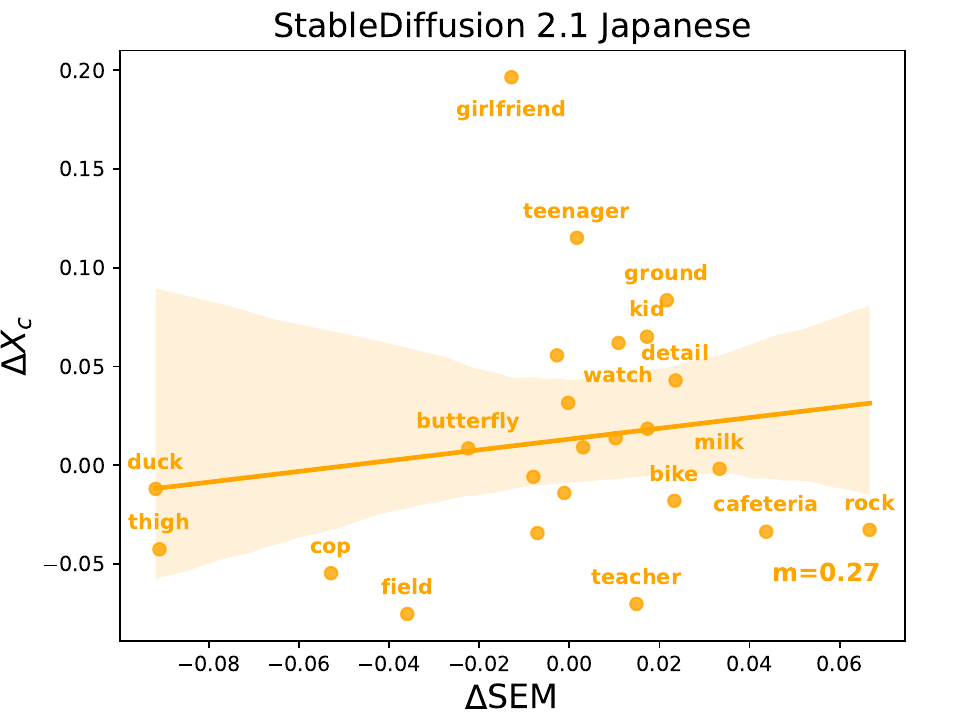}\includegraphics[width=0.33\linewidth]{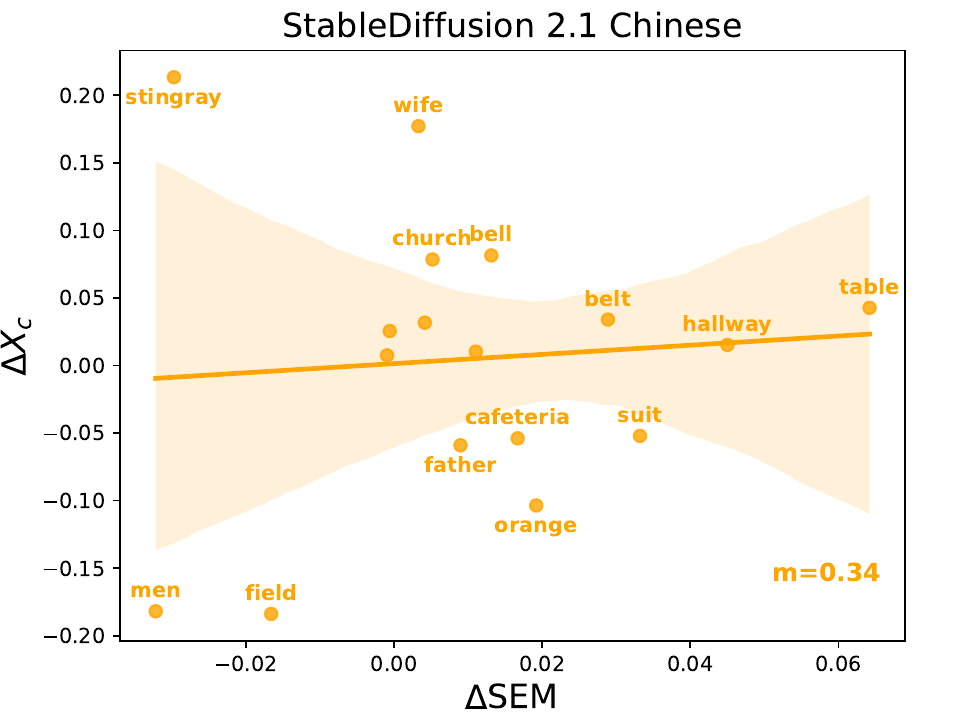}\includegraphics[width=0.33\linewidth]{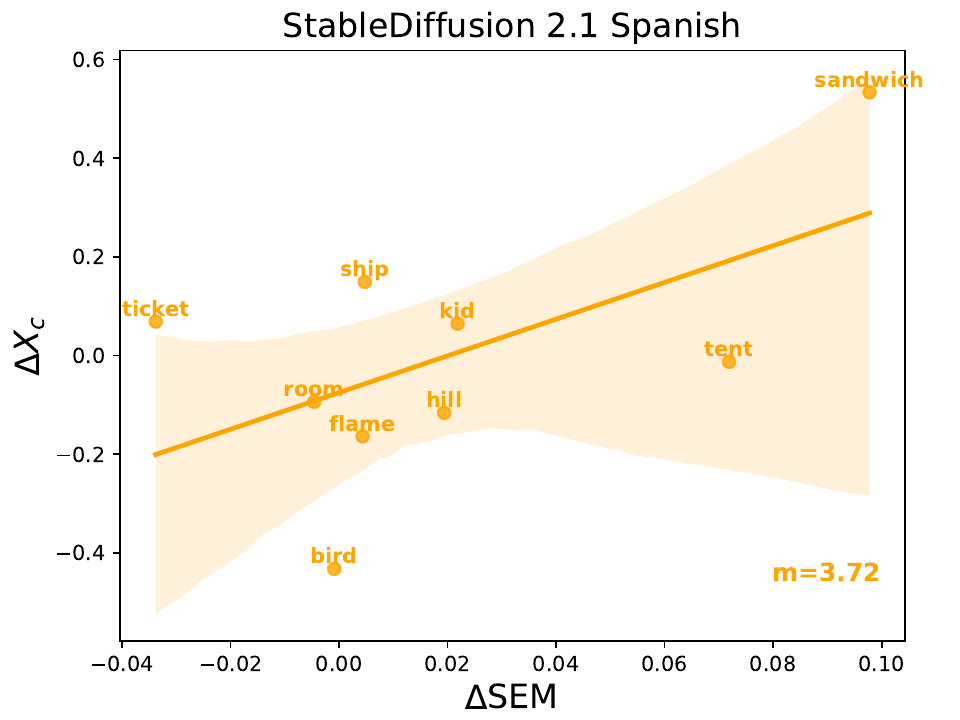} \\
\includegraphics[width=0.33\linewidth]{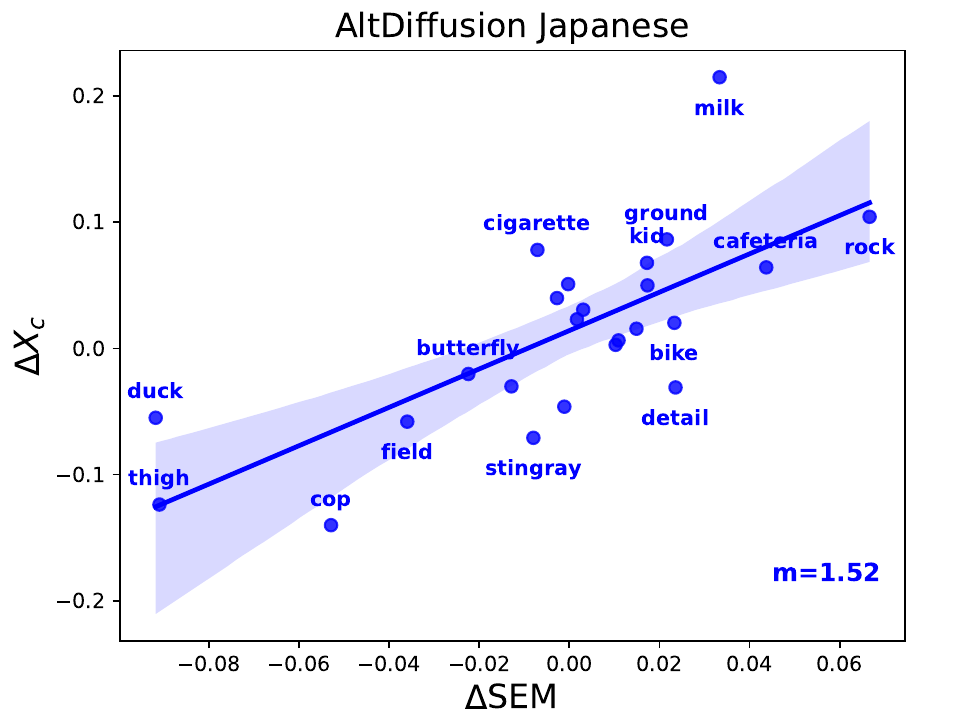}\includegraphics[width=0.33\linewidth]{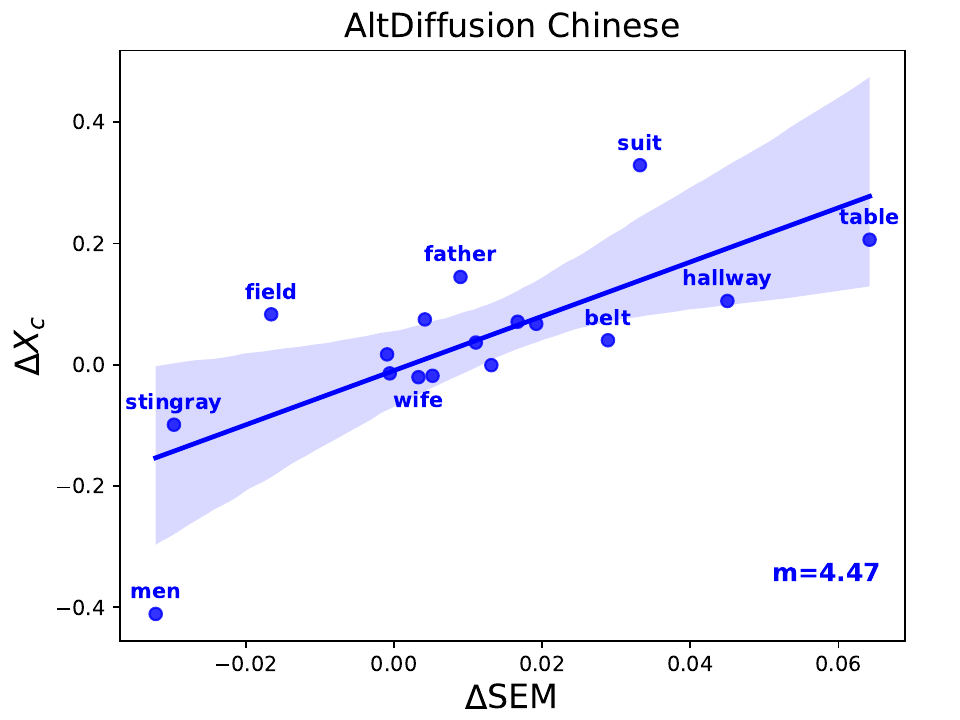}\includegraphics[width=0.33\linewidth]{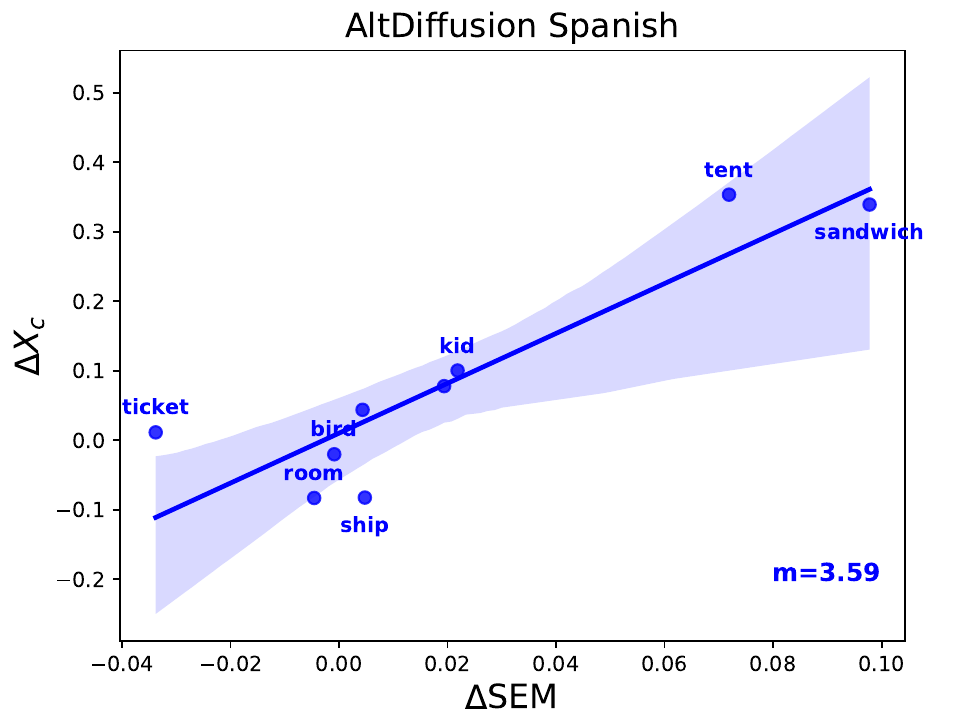}\\
\vspace{-7pt}{\scriptsize (a) Japanese \hspace{0.27\linewidth} (b) Chinese \hspace{0.27\linewidth} (c) Spanish}\vspace{-5pt}
\caption{Scatterplots showing the impact of the corrections to each concept in JA, ZH, and ES on the conceptwise improvement to the CCCL correctness score, $\Delta X_c$, as a function of $\Delta\mathrm{SEM}$. Slopes $m$ at bottom-right in \textbf{bold}.}\label{fig:scatcorr}
\vspace{-8pt}
\end{figure*}

\section{Results \& Analysis}

We generate output images using StableDiffusion 1.4, 2.0, 2.1 \cite{rombach2022high} and AltDiffusion \cite{chen2022altclip},
for all concepts corrected by our annotators in English, Spanish, Chinese, and Japanese, using both the original concept translations $c_\ell$ from CoCo-CroLa v1 \cite{saxon-wang-2023-multilingual} and the corrected translations $c'_\ell$. 
Model details are provided in Appendix \ref{sec:compdet}.

\autoref{fig:scatcorr} shows the relationship between $\Delta\mathrm{SEM}$ and $\Delta X_c$ for all corrected concepts for StableDiffusion 1.4, 2.0, 2.1, and AltDiffusion\footnote{Error margins are 95\% regression-fit confidence intervals.}. 
Note the pronounced, significant positive slope of the correlations between the two variables for AltDiffusion in all languages (\nth{4} row) and in Spanish for all models (third column). 
Here a positive slope means that \textit{higher-improvement} translation corrections (assessed by increased proximity to the English word in a shared embedding space) reliably correct the generated images more than the modest candidates. 

These same high-slope model/language pairs (eg., JA \& AltDiffusion) were found by \citet{saxon-wang-2023-multilingual} to be ``well-possessed'' (high average $X_c$ across correct concepts) in CoCo-CroLa v1. 
In other words, \textit{valid corrections only matter for languages a model already ``knows.''}

Correct Klingon is just as useless as incorrect Klingon to a non-Klingon model.

\autoref{tab:fitstats} (\autoref{sec:fullanalysis}) shows the same slopes $m$ with PCCs, $p$-values, and intercepts for the each model and language's $\Delta \mathrm{SEM}$ to $\Delta X_c$ relationship. The high-slope language/model pairs also tend to have higher PCC with more statistical significance. %

StableDiffusion 1.4 was trained on the primarily-Latin script \texttt{LAION-en-2b} \cite{schuhmann2021laion}, and thus lacks capabilities in non-Latin script languages JA and ZH. 
Consequently, there is no significant relationship between more semantically divergent corrections with high $\Delta\mathrm{SEM}$ and larger improvements to concept correctness $\Delta X_c$ for SD 1.4 on those languages.
Meanwhile, AltDiffusion---which conditions output images on the multilingual XLM-Roberta encoder \cite{conneau2020unsupervised}--benefits from all significant corrections in all languages with statistical significance.

\begin{figure}[!t]
\centering
\includegraphics[width=0.8\linewidth]{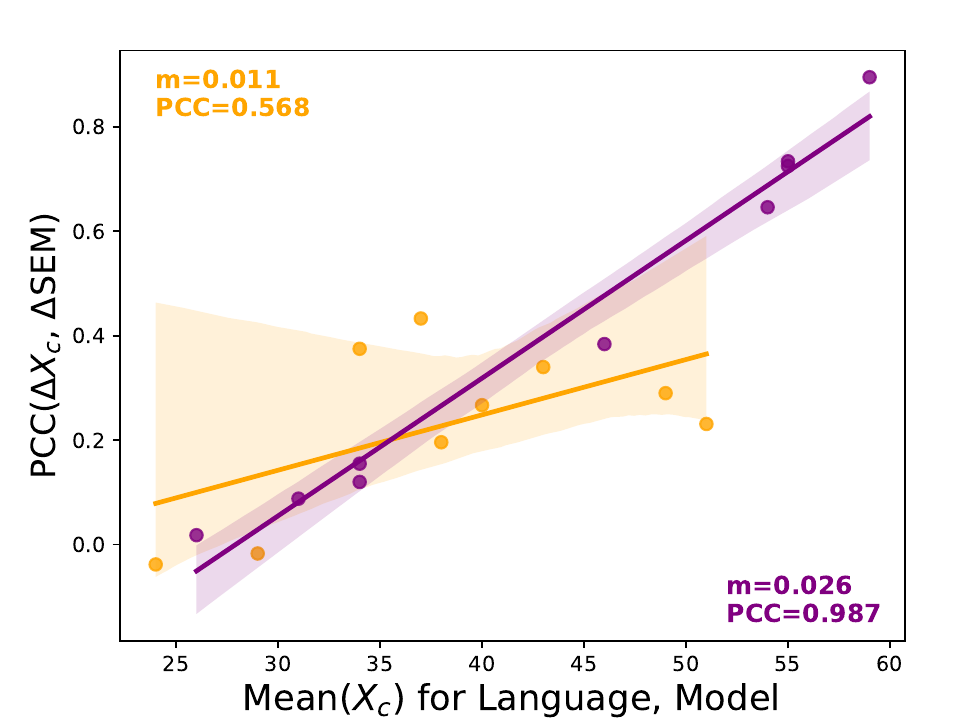}
\vspace{-1ex}
\caption{%
Languages with a high correlation between textual correction significance and image improvement (PCC) are more ``well-understood'' by the model ($X_c$), for both \textcolor{violet}{\textbf{real}}- and \textcolor{orange}{\textbf{pseudo}}-corrections.
}
\label{fig:pcc_xc}
\vspace{-3ex}
\end{figure}

\subsection{Pseudocorrection Experiment}

Unfortunately our ability to use the aforementioned corrections to confirm our hypothesis that 
\textit{T2I model language capability can be estimated from the impact of translation corrections on image-domain performance}
is hindered by the small quantity of correction candidates we found.
We bypass this problem with a \textit{pseudocorrection experiment}---simulating a larger set of corrections by generating artificial errors in the other CCCL languages.
We generate 10 synthetic erroneous \textit{pseudo-original translations}  for each concept in German, Indonesian, and Hebrew by randomly sampling the translations for other concepts within-language.
Each concept's ``correction'' is its original translation. 

For example, we assign the concept \textit{eye} the Indonesian word \textit{guru} (EN:teacher) as its pseudo-original. We then ``correct'' this word to \textit{mata}, the original correct translation, and assess $\Delta X_c$ and $\Delta\mathrm{SEM}$ with $c_{\ell_s}$:\textit{eye}, $c_\ell$:\textit{guru} and $c'_\ell$:\textit{mata}.

This gives us 1,930 $\Delta X_c$, $\Delta\mathrm{SEM}$ pairs for each language and model, with which we evaluate the same correlation relationship as before (plot in Appendix \autoref{fig:scatter_rand}). 
We report Pearson's correlation coefficient (PCC) for each of these pairs along with the average CCCL $X_c$ reported in \citet{saxon-wang-2023-multilingual} in \autoref{fig:pcc_xc}. The same relationship for real corrections holds for pseudocorrections, demonstrating that text-only multilingual semantic similarity features can predict the impact of a translation correction on the output image correctness.

\section{Discussion \& Conclusions}\label{sec:conclusions}

Our findings motivate important considerations for building future T2I semantic evaluations \cite{saharia2022photorealistic,Cho2022DallEval,huang2023t2icompbench}.

\paragraph{Subjectivity} A reliable T2I multilinguality assessment must report true knowledge failures---examples where a model fails to generate correct images of a concept, when it is correctly prompted to do so. 
Correct translations are required.

Unfortunately, choosing one ``correct translation'' is in inherently subjective task.
This study tackled this subjectivity by casting a wide net of error candidates, and testing their impact.
Consequential errors caused \textit{false negatives} where a concept to be erroneously marked as not possessed (\autoref{fig:figure1}).

CCCL's tangible concept constraint and corpus-based approach to finding concepts helps combat subjectivity \cite{saxon-wang-2023-multilingual}. In the tangible sense it's fair to say orange is correctly translated in Spanish to \textit{naranja} (the fruit) rather than \textit{anaranjado} (the adjective).

In prompting the T2I model we assume this tangible noun context is induced by using ``a picture of an $X$''-style prompts. While our results show this works, it is a model-specific phenomenon and future work should examine more prompt templates.

Future work grounded in prototype theory \cite{ando-etal-2002-extraction} may enable identification of culturally universal concepts for assessment.

\begin{figure*}[!t]
\centering
\includegraphics[width=\linewidth]{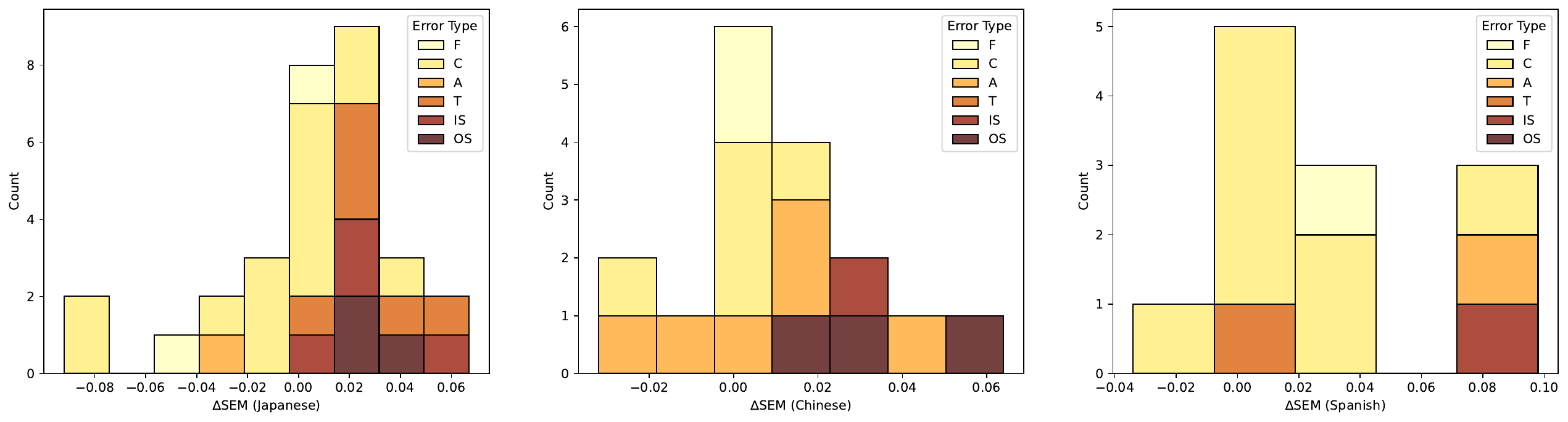}
\caption{%
Histograms for the error counts in JA, ZH, and ES vs $\Delta_{SEM}$, colored by error type. From lightest, they are F:\textit{formality}, C:\textit{commonality}, A:\textit{ambiguity}, T:\textit{transliteration}, IS:\textit{incoming sense error}, OS:\textit{outgoing sense error}. The error types are defined in \autoref{sec:errtype}. Severe error types will exhibit more rightward distributional mass.
}
\label{fig:hist}
\vspace{-12pt}
\end{figure*}

\paragraph{Need to assess Multiple Translations}
One challenge in multilinguality assessments is \textit{incoming duplicates}, where multiple ways of writing a translation really are equally correct. Our homograph errors have examples, such as cigarette in Japanese. \inlinejp{たばこ}, \inlinejp{タバコ}, and \inlinejp{煙草} are all translations of cigarette with identical reading, \textit{tabako}. %
Why should a metric of model-language capabilities only assess one correct translation rather than all?

More significant multiple translation problems arise in languages with gendered human-referent terms. For example, in Spanish \textit{maestro} refers to a male teacher, while \textit{maestra} a female one. Should a test of a model's Spanish knowledge of ``teacher'' as a concept test that both translations work equally well? 
CCCL v1 is incapable of assessing these attributes. 
Future benchmarks should contain this flexibility, so multiple incoming translations \cite{savoldi-etal-2021-gender} can be assessed for the same concept, while also tracing semantically-encoded secondary attributes such as gender between the source and test language.

\paragraph{Error Severity and Error Type}

\autoref{fig:hist} shows the distributions of error types for each language with respect to $\Delta_{SEM}$, our proxy for correction significance or error severity. Across all three languages, the \textit{sense errors} (OS and IS) are the most severe, while the formality and commonality errors are the least severe (defined in \autoref{sec:errtype}).

Our original estimated error rate (sum of all candidates per language) is a worst-case bound, the significant-to-evaluation-validity error rate is lower. Our impact and significance results show that some of our suggestions (mainly formality and commonality errors) may be more nitpick than correction.

Some concepts in CCCL are inherently erroneous due to intangibility. For example, \textit{history}, \textit{film}, and \textit{jump} are all present in v1 of CCCL, picked up for being high-frequency noun concepts across multiple languages in the corpora. There is no sensible prototypical way to generate images ``of'' those concepts. We removed these for CCCL v1.1; Future benchmarks should avoid including them.

\paragraph{Image-Image Metric Blind Spots}
We observed interesting borderline (potential false positive) cases where CoCo-CroLa scored mistranslated concepts as possessed. 
For example, \textit{bike} in Japanese. \autoref{fig:figure1} shows that under the erroneous translation, AltDiffusion generates pictures of \textit{motorcycles} rather than \textit{bicycles} as it does in English. However, $X_c$ doesn't actually change much under this correction as shown in \autoref{fig:scatcorr} \& \autoref{tab: mistranslation-examples-all}.
The CLIP similarity score in CCCL is blind to the difference between a bicycle and motorcycle. 
Mistranslations where visual structural similarity is present are sometimes invisible to the image metrics.

\paragraph{Tangible object translation as an MT domain} Single word concepts are not central to the distribution of machine translation training data. By providing the individual English tangible nouns as input we may expect an unreasonable amount of implicit commonsense reasoning from commercial MT systems---the correct sense out of many had to be selected for success.
Furthermore, the use of the BabelNet knowledge graph as a consensus mechanism reinforced some sense errors. For example, the \textit{rock} sense error for JA (music genre rather than physical object, \autoref{tab: mistranslation-examples-all}) was also present in Hebrew, probably due to shared edges in the knowledge graph.
Given previous interest in assessing the performance of MT translation in diverse domains \cite{irvine-etal-2013-measuring}, we think both the word-level translation of concepts under domain constraints without context (as we tried to do in CCCL previously) and treating input prompts for T2I systems (ie, captions) \cite{hitschler2016multimodal, singh2021multiple} as a target domain for MT evaluation would be interesting and useful future directions.

Future benchmarks should leverage context with sentences as input to MT (eg, ``watch for falling rocks'') rather than the decontextualized concept words alone to improve robustness. LLMs could generate diverse English sentence examples, and could potentially also extract the final concept translations out of the multiple sentence translations.

\section*{Limitations}

Trivially, human annotators for every language would remove false-negative mistranslations from future benchmarks, but %
there's a trade-off between easy scalability and certainty of correctness.

Our work incorporates human efforts of both native and proficient but non-native language speakers to propose and resolve translation error candidates caused by the machine translation pipeline in the original CoCo-CroLa benchmark. This could potentially bring human biases into the nuance of factors such as words' choices, introducing less culturally neural expressions as a result.  

The assumption of \textit{translatability} that underlies CCCL in general is a challenge.
As a practical use-based test of functional fairness, using heuristics and only common everyday objects that can be reasonably assumed universal is acceptable, but more linguistic and even philosophical work is needed to really motivate fairness across languages and cultures when underlying assumptions differ.

\section*{Acknowledgements}

Thanks to our December 2023 ARR reviewers and ACs, particularly Yx9V for thoughtful and detailed reviewing and conversation, and many useful suggestions. 
Thank you Alfonso Amayuelas for feedback on ES candidates.
This work was supported in part by the National Science Foundation Graduate Research Fellowship under Grant No. 1650114, and CAREER Award under Grant No. 2048122.

\bibliography{anthology,custom}

\begin{thebibliography}{26}
\expandafter\ifx\csname natexlab\endcsname\relax\def\natexlab#1{#1}\fi

\bibitem[{Agrawal et~al.(2018)Agrawal, Batra, Parikh, and Kembhavi}]{vqa-cp}
Aishwarya Agrawal, Dhruv Batra, Devi Parikh, and Aniruddha Kembhavi. 2018.
\newblock Don't just assume; look and answer: Overcoming priors for visual question answering.
\newblock In \emph{Proceedings of the IEEE Conference on Computer Vision and Pattern Recognition}, pages 4971--4980.

\bibitem[{Ando et~al.(2002)Ando, Okamoto, and Ishizaki}]{ando-etal-2002-extraction}
Maya Ando, Jun Okamoto, and Shun Ishizaki. 2002.
\newblock \href {http://www.lrec-conf.org/proceedings/lrec2002/pdf/270.pdf} {Extraction of associative attributes from nouns and quantitative expression of prototype concept}.
\newblock In \emph{Proceedings of the Third International Conference on Language Resources and Evaluation ({LREC}{'}02)}, Las Palmas, Canary Islands - Spain. European Language Resources Association (ELRA).

\bibitem[{Antol et~al.(2015)Antol, Agrawal, Lu, Mitchell, Batra, Zitnick, and Parikh}]{vqav1}
Stanislaw Antol, Aishwarya Agrawal, Jiasen Lu, Margaret Mitchell, Dhruv Batra, C~Lawrence Zitnick, and Devi Parikh. 2015.
\newblock Vqa: Visual question answering.
\newblock In \emph{Proceedings of the IEEE international conference on computer vision}, pages 2425--2433.

\bibitem[{Chen et~al.(2022)Chen, Liu, Zhang, Ye, Yang, and Wu}]{chen2022altclip}
Zhongzhi Chen, Guang Liu, Bo-Wen Zhang, Fulong Ye, Qinghong Yang, and Ledell Wu. 2022.
\newblock \href {https://arxiv.org/abs/2211.06679} {Altclip: Altering the language encoder in clip for extended language capabilities}.
\newblock \emph{ArXiv preprint}, abs/2211.06679.

\bibitem[{Cho et~al.(2022)Cho, Zala, and Bansal}]{Cho2022DallEval}
Jaemin Cho, Abhay Zala, and Mohit Bansal. 2022.
\newblock \href {http://arxiv.org/abs/2202.04053} {Dall-eval: Probing the reasoning skills and social biases of text-to-image generative transformers}.

\bibitem[{Conneau et~al.(2020)Conneau, Khandelwal, Goyal, Chaudhary, Wenzek, Guzm{\'a}n, Grave, Ott, Zettlemoyer, and Stoyanov}]{conneau2020unsupervised}
Alexis Conneau, Kartikay Khandelwal, Naman Goyal, Vishrav Chaudhary, Guillaume Wenzek, Francisco Guzm{\'a}n, Edouard Grave, Myle Ott, Luke Zettlemoyer, and Veselin Stoyanov. 2020.
\newblock \href {https://doi.org/10.18653/v1/2020.acl-main.747} {Unsupervised cross-lingual representation learning at scale}.
\newblock In \emph{ACL 2020}, pages 8440--8451, Online. Association for Computational Linguistics.

\bibitem[{Cui et~al.(2021)Cui, Khandelwal, Artzi, Snavely, and Averbuch-Elor}]{whoswaldo}
Yuqing Cui, Apoorv Khandelwal, Yoav Artzi, Noah Snavely, and Hadar Averbuch-Elor. 2021.
\newblock Who's waldo? linking people across text and images.
\newblock In \emph{Proceedings of the IEEE/CVF International Conference on Computer Vision (ICCV)}, pages 1374--1384.

\bibitem[{Hitschler et~al.(2016)Hitschler, Schamoni, and Riezler}]{hitschler2016multimodal}
Julian Hitschler, Shigehiko Schamoni, and Stefan Riezler. 2016.
\newblock Multimodal pivots for image caption translation.
\newblock In \emph{Proceedings of the 54th Annual Meeting of the Association for Computational Linguistics (Volume 1: Long Papers)}. Association for Computational Linguistics.

\bibitem[{Ho et~al.(2023)Ho, Sharma, Chang, Saxon, Levy, Lu, and Wang}]{ho2023wikiwhy}
Matthew Ho, Aditya Sharma, Justin Chang, Michael Saxon, Sharon Levy, Yujie Lu, and William~Yang Wang. 2023.
\newblock \href {https://openreview.net/forum?id=vaxnu-Utr4l} {Wikiwhy: Answering and explaining cause-and-effect questions}.
\newblock In \emph{The Eleventh International Conference on Learning Representations}.

\bibitem[{Huang et~al.(2023)Huang, Sun, Xie, Li, and Liu}]{huang2023t2icompbench}
Kaiyi Huang, Kaiyue Sun, Enze Xie, Zhenguo Li, and Xihui Liu. 2023.
\newblock T2i-compbench: A comprehensive benchmark for open-world compositional text-to-image generation.
\newblock \emph{arXiv preprint arXiv:2307.06350}.

\bibitem[{Huang et~al.(2024)Huang, Sun, Xie, Li, and Liu}]{huang2024t2i}
Kaiyi Huang, Kaiyue Sun, Enze Xie, Zhenguo Li, and Xihui Liu. 2024.
\newblock T2i-compbench: A comprehensive benchmark for open-world compositional text-to-image generation.
\newblock \emph{Advances in Neural Information Processing Systems}, 36.

\bibitem[{Irvine et~al.(2013)Irvine, Morgan, Carpuat, Daum{\'e}~III, and Munteanu}]{irvine-etal-2013-measuring}
Ann Irvine, John Morgan, Marine Carpuat, Hal Daum{\'e}~III, and Dragos Munteanu. 2013.
\newblock \href {https://doi.org/10.1162/tacl_a_00239} {Measuring machine translation errors in new domains}.
\newblock \emph{Transactions of the Association for Computational Linguistics}, 1:429--440.

\bibitem[{Luo et~al.(2022)Luo, Banerjee, Gokhale, Yang, and Baral}]{tofindwaldo}
Yiran Luo, Pratyay Banerjee, Tejas Gokhale, Yezhou Yang, and Chitta Baral. 2022.
\newblock \href {https://doi.org/10.18653/v1/2022.acl-short.39} {To find waldo you need contextual cues: Debiasing who’s waldo}.
\newblock In \emph{Proceedings of the 60th Annual Meeting of the Association for Computational Linguistics (Volume 2: Short Papers)}, page 355–361, Dublin, Ireland. Association for Computational Linguistics.

\bibitem[{Navigli and Ponzetto(2010)}]{navigli2010babelnet}
Roberto Navigli and Simone~Paolo Ponzetto. 2010.
\newblock \href {https://aclanthology.org/P10-1023} {{B}abel{N}et: Building a very large multilingual semantic network}.
\newblock In \emph{Proceedings of the 48th Annual Meeting of the Association for Computational Linguistics}, pages 216--225, Uppsala, Sweden. Association for Computational Linguistics.

\bibitem[{Patel et~al.(2024)Patel, Gokhale, Baral, and Yang}]{patel2023conceptbed}
Maitreya Patel, Tejas Gokhale, Chitta Baral, and Yezhou Yang. 2024.
\newblock Conceptbed: Evaluating concept learning abilities of text-to-image diffusion models.

\bibitem[{Radford et~al.(2021)Radford, Kim, Hallacy, Ramesh, Goh, Agarwal, Sastry, Askell, Mishkin, Clark, Krueger, and Sutskever}]{radford2021learning}
Alec Radford, Jong~Wook Kim, Chris Hallacy, Aditya Ramesh, Gabriel Goh, Sandhini Agarwal, Girish Sastry, Amanda Askell, Pamela Mishkin, Jack Clark, Gretchen Krueger, and Ilya Sutskever. 2021.
\newblock \href {http://proceedings.mlr.press/v139/radford21a.html} {Learning transferable visual models from natural language supervision}.
\newblock In \emph{ICML 2021}, volume 139 of \emph{Proceedings of Machine Learning Research}, pages 8748--8763. {PMLR}.

\bibitem[{Reimers and Gurevych(2019)}]{reimers-2019-sentence-bert}
Nils Reimers and Iryna Gurevych. 2019.
\newblock \href {http://arxiv.org/abs/1908.10084} {Sentence-bert: Sentence embeddings using siamese bert-networks}.
\newblock In \emph{Proceedings of the 2019 Conference on Empirical Methods in Natural Language Processing}. Association for Computational Linguistics.

\bibitem[{Rombach et~al.(2022)Rombach, Blattmann, Lorenz, Esser, and Ommer}]{rombach2022high}
Robin Rombach, Andreas Blattmann, Dominik Lorenz, Patrick Esser, and Bj{\"o}rn Ommer. 2022.
\newblock High-resolution image synthesis with latent diffusion models.
\newblock In \emph{CVPR 2022}, pages 10684--10695.

\bibitem[{Saharia et~al.(2022)Saharia, Chan, Saxena, Li, Whang, Denton, Ghasemipour, Gontijo~Lopes, Karagol~Ayan, Salimans, Ho, Fleet, and Norouzi}]{saharia2022photorealistic}
Chitwan Saharia, William Chan, Saurabh Saxena, Lala Li, Jay Whang, Emily~L Denton, Kamyar Ghasemipour, Raphael Gontijo~Lopes, Burcu Karagol~Ayan, Tim Salimans, Jonathan Ho, David~J Fleet, and Mohammad Norouzi. 2022.
\newblock \href {https://proceedings.neurips.cc/paper_files/paper/2022/file/ec795aeadae0b7d230fa35cbaf04c041-Paper-Conference.pdf} {Photorealistic text-to-image diffusion models with deep language understanding}.
\newblock 35:36479--36494.

\bibitem[{Savoldi et~al.(2021)Savoldi, Gaido, Bentivogli, Negri, and Turchi}]{savoldi-etal-2021-gender}
Beatrice Savoldi, Marco Gaido, Luisa Bentivogli, Matteo Negri, and Marco Turchi. 2021.
\newblock \href {https://doi.org/10.1162/tacl_a_00401} {Gender bias in machine translation}.
\newblock \emph{Transactions of the Association for Computational Linguistics}, 9:845--874.

\bibitem[{Saxon and Wang(2023)}]{saxon-wang-2023-multilingual}
Michael Saxon and William~Yang Wang. 2023.
\newblock \href {https://doi.org/10.18653/v1/2023.acl-long.266} {Multilingual conceptual coverage in text-to-image models}.
\newblock In \emph{Proceedings of the 61st Annual Meeting of the Association for Computational Linguistics (Volume 1: Long Papers)}, pages 4831--4848, Toronto, Canada. Association for Computational Linguistics.

\bibitem[{Saxon et~al.(2023)Saxon, Wang, Xu, and Wang}]{saxon-etal-2023-peco}
Michael Saxon, Xinyi Wang, Wenda Xu, and William~Yang Wang. 2023.
\newblock \href {https://doi.org/10.18653/v1/2023.eacl-main.223} {{PECO}: Examining single sentence label leakage in natural language inference datasets through progressive evaluation of cluster outliers}.
\newblock In \emph{Proceedings of the 17th Conference of the European Chapter of the Association for Computational Linguistics}, pages 3061--3074, Dubrovnik, Croatia. Association for Computational Linguistics.

\bibitem[{Schuhmann et~al.(2021)Schuhmann, Vencu, Beaumont, Kaczmarczyk, Mullis, Katta, Coombes, Jitsev, and Komatsuzaki}]{schuhmann2021laion}
Christoph Schuhmann, Richard Vencu, Romain Beaumont, Robert Kaczmarczyk, Clayton Mullis, Aarush Katta, Theo Coombes, Jenia Jitsev, and Aran Komatsuzaki. 2021.
\newblock \href {https://arxiv.org/abs/2111.02114} {Laion-400m: Open dataset of clip-filtered 400 million image-text pairs}.
\newblock \emph{ArXiv preprint}, abs/2111.02114.

\bibitem[{Singh et~al.(2021)Singh, Meetei, Singh, and Bandyopadhyay}]{singh2021multiple}
Salam~Michael Singh, Loitongbam~Sanayai Meetei, Thoudam~Doren Singh, and Sivaji Bandyopadhyay. 2021.
\newblock Multiple captions embellished multilingual multi-modal neural machine translation.
\newblock In \emph{Proceedings of the First Workshop on Multimodal Machine Translation for Low Resource Languages (MMTLRL 2021)}, pages 2--11.

\bibitem[{Ye and Kovashka(2021)}]{vcr-shortcut}
Keren Ye and Adriana Kovashka. 2021.
\newblock A case study of the shortcut effects in visual commonsense reasoning.
\newblock In \emph{Proceedings of the AAAI Conference on Artificial Intelligence}, volume~35, pages 3181--3189.

\bibitem[{Zellers et~al.(2019)Zellers, Bisk, Farhadi, and Choi}]{vcr}
Rowan Zellers, Yonatan Bisk, Ali Farhadi, and Yejin Choi. 2019.
\newblock From recognition to cognition: Visual commonsense reasoning.
\newblock In \emph{Proceedings of the IEEE/CVF Conference on Computer Vision and Pattern Recognition}, pages 6720--6731.

\end{thebibliography}

\appendix

\section{Appendix}

\label{sec:appendix}

\subsection{Contribution Statement}

YL produced the Chinese and Japanese translation error candidates and the overall EC taxonomy. MS produced the Spanish candidates and checked the Japanese candidates. YL evaluated $\Delta\mathrm{SEM}$, MS generated the before/after images and evaluated $X_c$ and $\Delta X_c$. YL produced diagrams and MS graphs.

\subsubsection{Human Annotation Details}\label{sec:procedure}

MS and YL produced the initial list of candidate errors and corrections.
MS is a native speaker of English and literate second language speaker of Spanish and Japanese. YL is a native speaker of Chinese, professionally proficient speaker of English, and a literate proficient speaker of Japanese, with experience in literary translation and textual localization between English, Chinese, and Japanese. 

Each annotator first read through the list of their languages (ES/JA and ZH/JA respectively) for about 10 minutes and marked every translation (\textit{error candidate}) that appeared incorrect with a preliminary correction. They then verified the annotations using bilingual English-\{Spanish, Japanese, Chinese\} resources and consultation with native speakers where relevant as detailed below.

MS checked Spanish corrections using Spanish-language example usage notes provided in the Spanish \href{https://www.wordreference.com/es/translation.asp}{wordreference.com} dictionary, and consultation with a native speaker. MS's JA error candidates were a subset of YL's. YL also took references from language standard dictionaries used by native speakers---for Chinese \textit{Xiandai Hanyu Cidian} and for Japanese \textit{Shin Meikai Kokugo Jiten}.

\subsection{Additional Resource Information}\label{sec:resdet}

\paragraph{Intended Use, License and Terms} We release our corrections as a v1.1 revision to the CoCo-CroLa benchmark \cite{saxon-wang-2023-multilingual} intended to evaluate the performance of text-to-image models. It inherits v1's license and terms.

\paragraph{Offensive Content} Some of the erroneous translations we found can lead to offensive images, e.g. the original JA translation for \textit{milk} in also means ``breast.''

\subsection{Error candidate typology}\label{sec:errtype}

\noindent \textbf{Commonality (C). } When a selected translated term doesn't appear to reflect the most common, colloquial, contemporary, or ``natural'' way that native speakers of the language would use in reference to the concept in a photograph or conversation. For example, in Chinese ``\inlinezh{瓶子}'' is a more conversational and contemporary way of writing \textit{bottle} than ``\inlinezh{瓶},'' which reads literary and archaic.

\noindent \textbf{Outgoing Sense Error. (OS) }The translated term picks an alternative (and often less tangible) sense from the source concept. For example, the original Chinese translation for \textit{Table} diverges to the sense of `spreadsheet, tabular', instead of the presumptive home furniture item. 

\noindent \textbf{Incoming Sense Error. (IS) }The translated term, while aligned to the correct source concept sense, picks a phrasing for which other senses in the target language exist that the annotators expect will confound model behavior, where another (often more common) disambiguated translation also exists.
For example, the original Spanish translation for \textit{tent} is given as \textit{tienda} alone, which can also mean `store, shop', in addition to `a tent,' whereas the corrected translation \textit{tienda de acampar} refers to a camping tent alone.

\noindent \textbf{Ambiguity (A).} The translated term introduces a word with multiple meanings from the unambiguous source concept. For example, the Japanese translation for \textit{Milk} originally uses a single character that can mean any kind of animal or human milk, or even the organ of the breast.  

\noindent \textbf{Formality. (F)}The translated term uses an expression in an improper formality. For example, the original Chinese translation for \textit{Father} is only heard in casual conversations. 

\noindent \textbf{Transliteration (T). }When one of the above errors occurs with 
. For example, the transliteration of \textit{Rock} in Japanese is commonly related to `Rock Music', rather than stones found in nature.

\subsection{Computational Experiments Details}\label{sec:compdet}

\paragraph{Dataset Statistics} CCCL contains 193 multilingual concepts written in 7 languages. We have also modified 50 of these in ES, ZH, or JA with verified translations by human annotators.
\paragraph{Models Employed} See \autoref{tab:models}.

\begin{table}[h!]
\small
    \centering
    \resizebox{\linewidth}{!}{
    \begin{tabular}{lrl}
        \toprule
        Model &  \# Param & Repository (\texttt{huggingface.co/...}) \\
        \midrule
        StableDiffusion 1.4 & ~860M & \href{https://huggingface.co/CompVis/stable-diffusion-v1-4}{\texttt{CompVis/stable-diffusion-v1-4}} \\
        StableDiffusion 2 & NA & \href{https://github.com/Stability-AI/stablediffusion}{\texttt{stabilityai/stable-diffusion-2}} \\
        StableDiffusion 2.1 & NA & \href{https://github.com/Stability-AI/stablediffusion}{\texttt{stabilityai/stable-diffusion-2}} \\
        AltDiffusion m9 &~1.7B& \href{https://huggingface.co/BAAI/AltDiffusion-m9/tree/main}{\texttt{BAAI/AltDiffusion-m9}} \\
        \bottomrule
    \end{tabular}
    }
    \vspace{-5pt}
    \caption{The set of text-to-image models we evaluated with (Table adapted from \cite{saxon-wang-2023-multilingual}.} %
    \vspace{-2ex}
    \label{tab:models}
\end{table}

\paragraph{Experimental Setup} %
We generated 9 images for each (language, model, concept) triple and evaluated $X_C$ using identical methods and code%
as described in CCCL \cite{saxon-wang-2023-multilingual}.

\subsection{Full Analysis Numbers}\label{sec:fullanalysis}

\begin{table}[h!]\centering
\small
\begin{tabular}{llrrllrrllrr}
\toprule
Model & Language & PCC & $p$ & $m$ & $b$ \\\midrule
SD1-4 & Japanese & 0.120 & 0.577 & 0.437 & 0.049 \\
SD1-4 & Chinese & 0.018 & 0.944 & 0.051 & -0.011 \\
SD1-4 & Spanish & 0.384 & 0.307 & 1.877 & -0.064 \\
SD2 & Japanese & 0.088 & 0.684 & 0.155 & 0.020 \\
SD2 & Chinese & 0.155 & 0.554 & 0.608 & 0.000 \\
SD2 & Spanish & 0.646 & 0.060 & 3.891 & -0.067 \\
AD & Japanese & 0.734 & 0.000 & 1.519 & 0.014 \\
AD & Chinese & 0.725 & 0.001 & 4.472 & -0.010 \\
AD & Spanish & 0.895 & 0.001 & 3.588 & 0.010 \\
SD2-1 & Japanese & 0.162 & 0.448 & 0.272 & 0.013 \\
SD2-1 & Chinese & 0.078 & 0.765 & 0.340 & 0.001 \\
SD2-1 & Spanish & 0.574 & 0.106 & 3.722 & -0.075 \\
\bottomrule
 \end{tabular}
\caption{
Stats for Pearson correlation and linear best fit between $\Delta\mathrm{SEM}$ and $\Delta X_c$ for each model and language. 
$p$ represents the $p$-value for the PCC, $m$ and $b$ the slope and intercept for the best-fit line. 
} \label{tab:fitstats}
\vspace{-12pt}
\end{table}

\subsection{Further Related Work}

ConceptBed \cite{patel2023conceptbed} evaluates monolingual concept-level knowledge in T2I, and its concept inventory could extend and improve CCCL's.
T2I-CompBench assesses compositionality in T2I \cite{huang2024t2i}, leveraging VQA and image segmentation.
Assessment model weaknesses, such as \citet{vqa-cp}'s VQA spurious correlations \citep{vqav1} remain a challenge.

Other benchmarks in vision-and-language also require correction and improvement. \citet{tofindwaldo} found and filtered unsolvable cases in \textit{Who's Waldo} \citep{whoswaldo}.
\citet{vcr-shortcut} exploit repeated texts in QA pairs on \textit{VCR} \citep{vcr}. While manual techniques can find and clean these errors, automated approaches would be preferable, such as the PECO method \cite{saxon-etal-2023-peco} for finding model-used shortcuts in NLI.
Semi-human-in-the-loop approaches \cite{ho2023wikiwhy} may improve the sourcing and cleaning of future CCCL versions.

\begin{figure*}[!t]
\centering
\includegraphics[width=\linewidth]{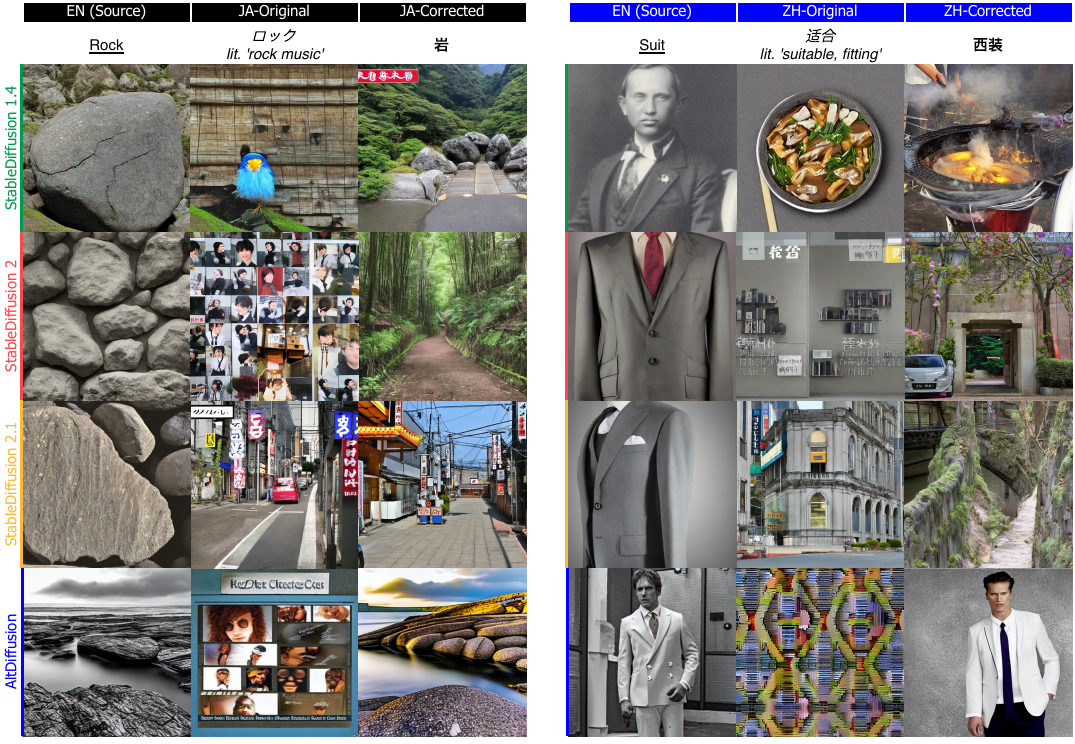}
\includegraphics[width=\linewidth]{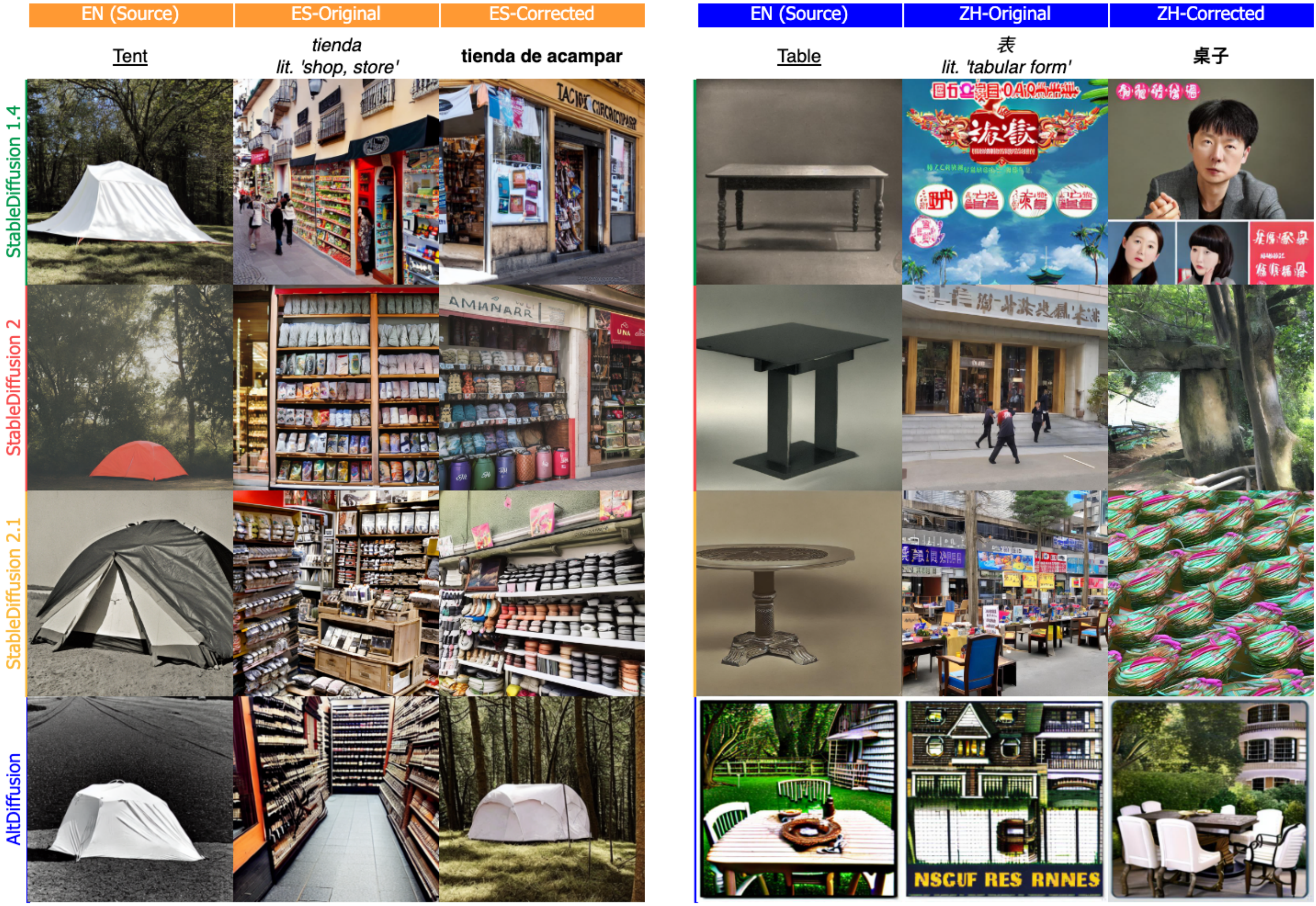}
\caption{Qualitative examples of selected mistranslated concepts found in Coco-CroLa generated by AltDiffusion and multiple versions of Stable Diffusion - \textbf{Top left: }``Rock'' in Japanese, \textbf{Top right: }``Suit'' in Chinese, \textbf{Bottom left: }``Tent'' in Spanish, \textbf{Bottom right: }``Table'' in Chinese. Noticeably, we observe that T2I models such as Stable Diffusion 2 do not benefit from correcting the translations, as their outputs in the aforementioned languages remain irrelevant similarly to using random prompts. }
\label{fig:re-eval}
\end{figure*}

\newpage

\begin{figure*}[!h]
\centering
\includegraphics[width=0.33\linewidth]{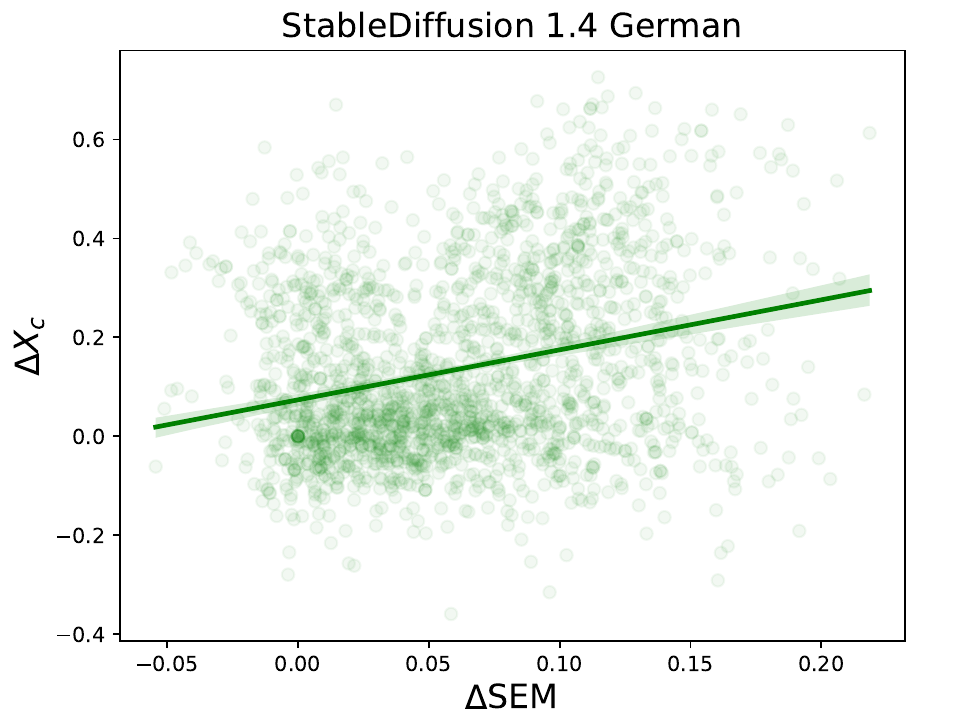}\includegraphics[width=0.33\linewidth]{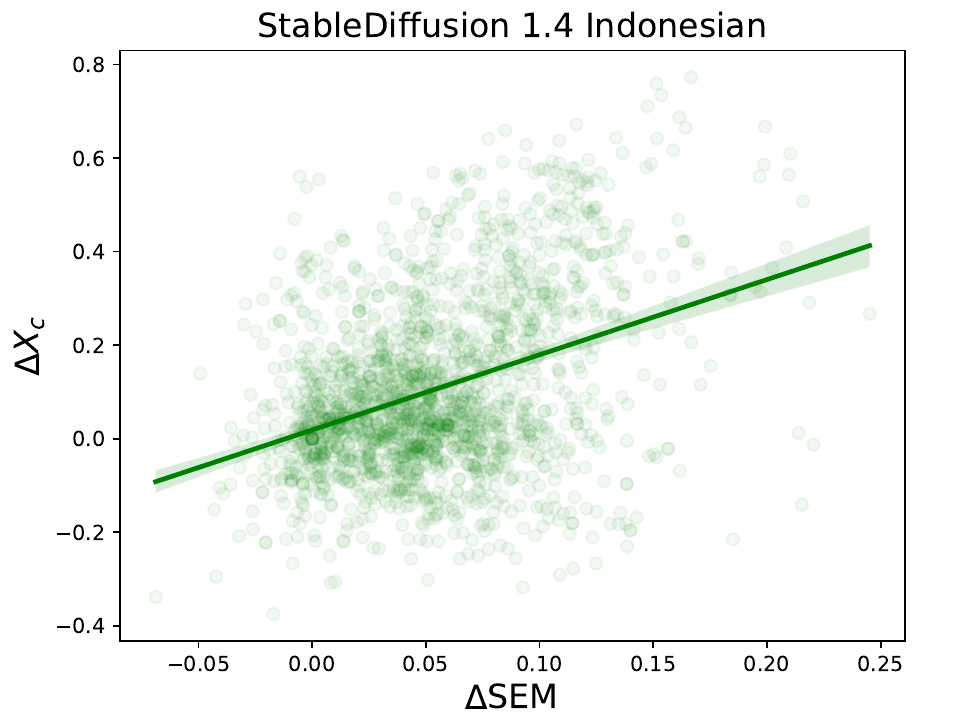}\includegraphics[width=0.33\linewidth]{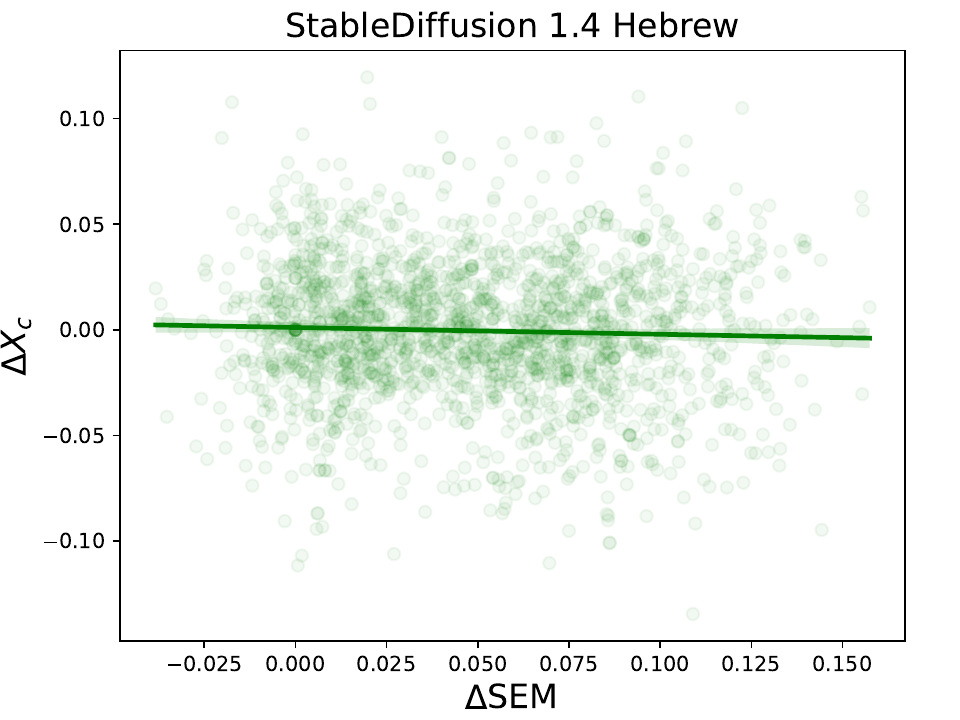} \\
\includegraphics[width=0.33\linewidth]{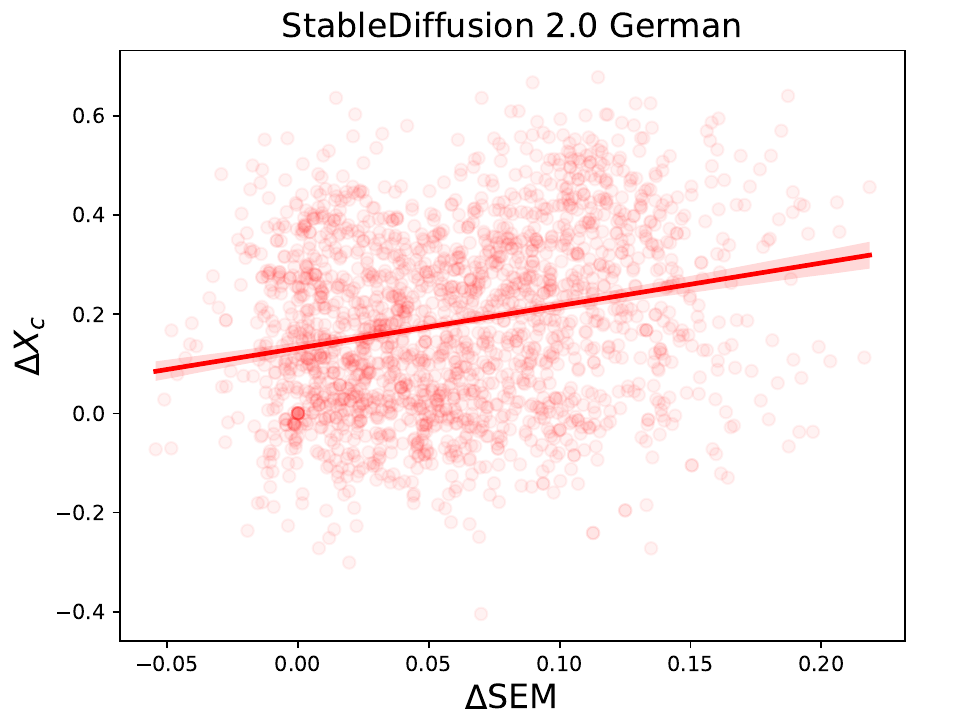}\includegraphics[width=0.33\linewidth]{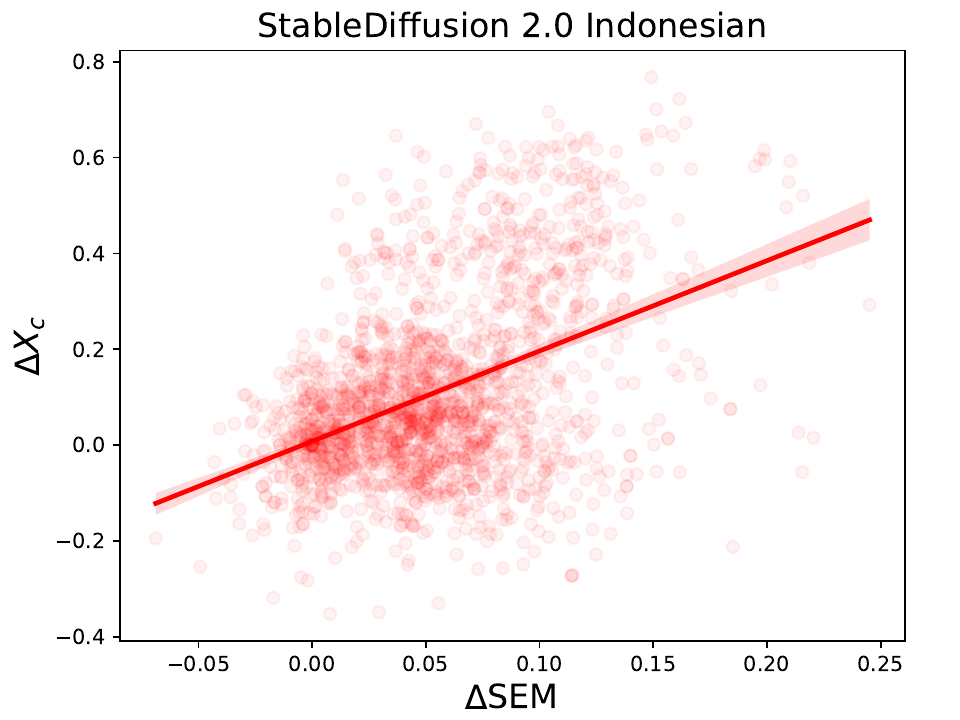}\includegraphics[width=0.33\linewidth]{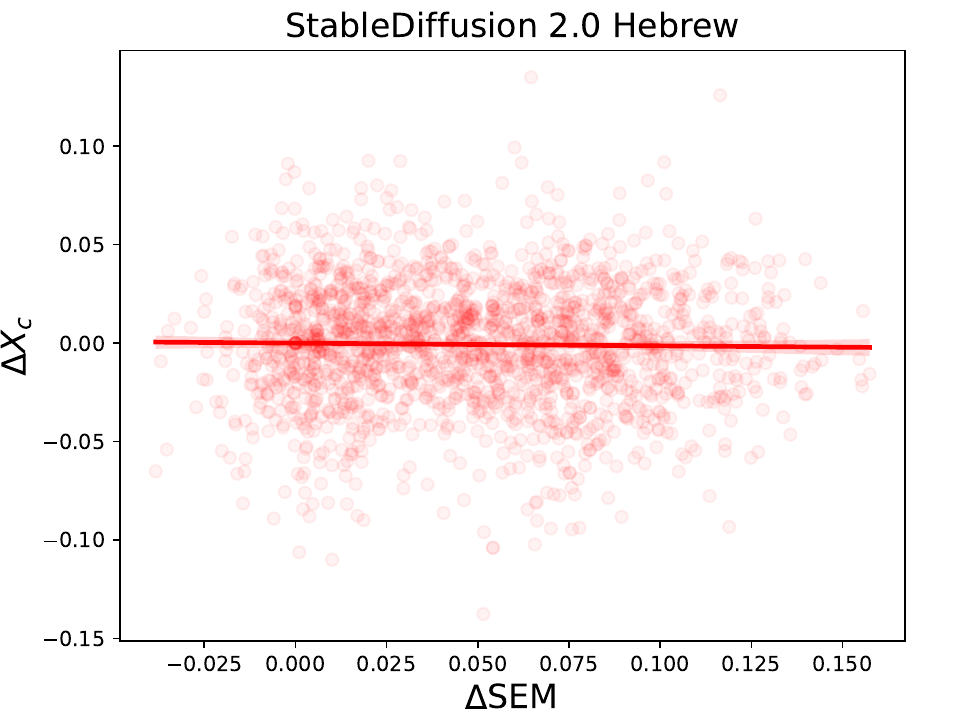} 
\includegraphics[width=0.33\linewidth]{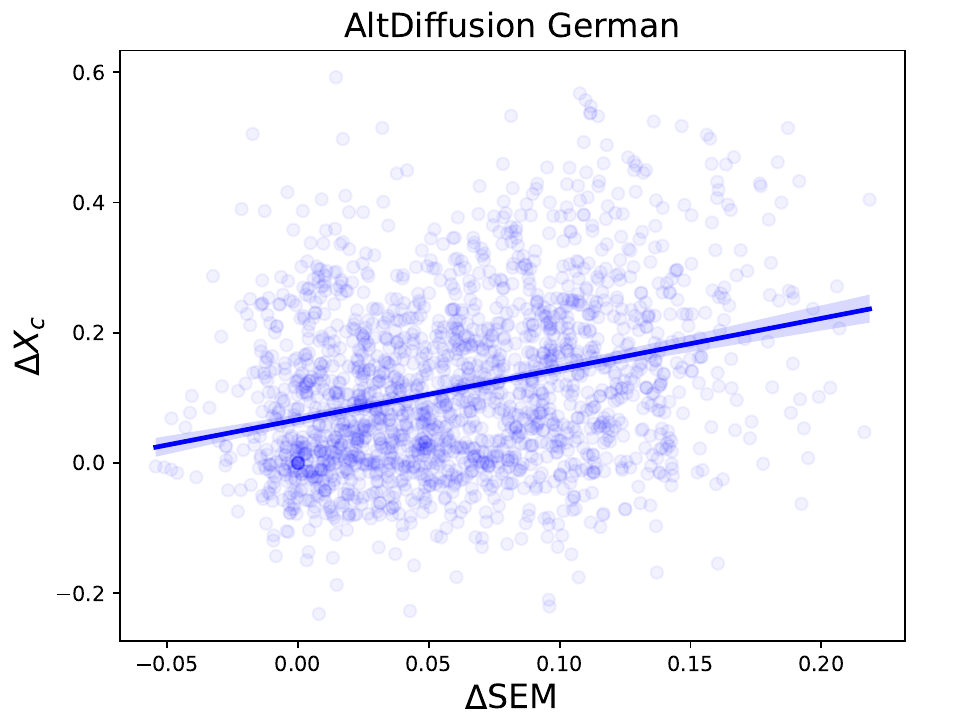}\includegraphics[width=0.33\linewidth]{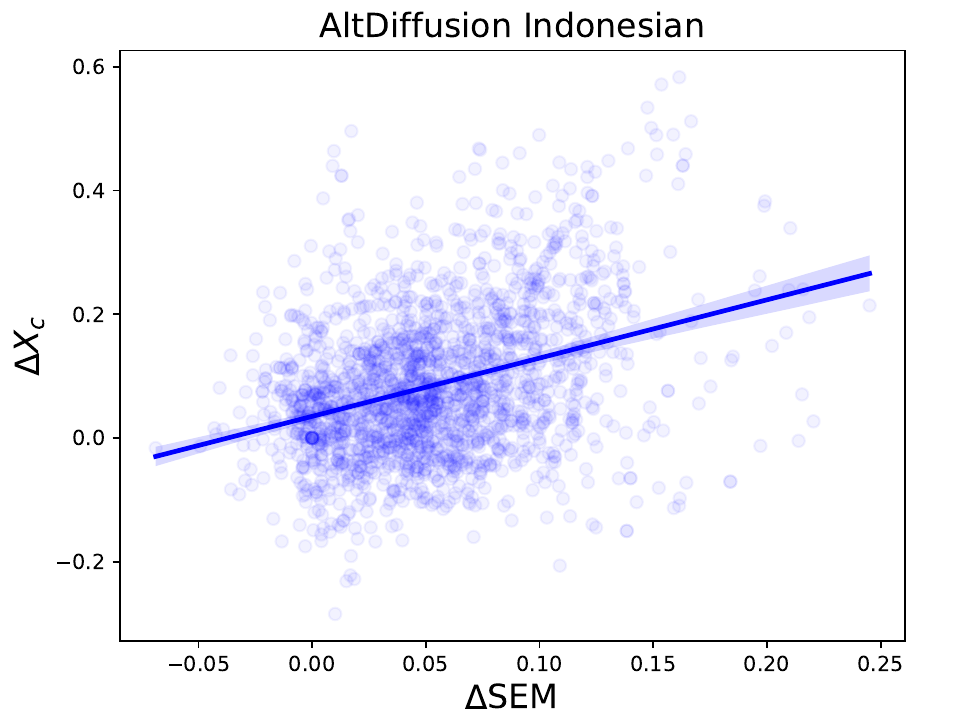}\includegraphics[width=0.33\linewidth]{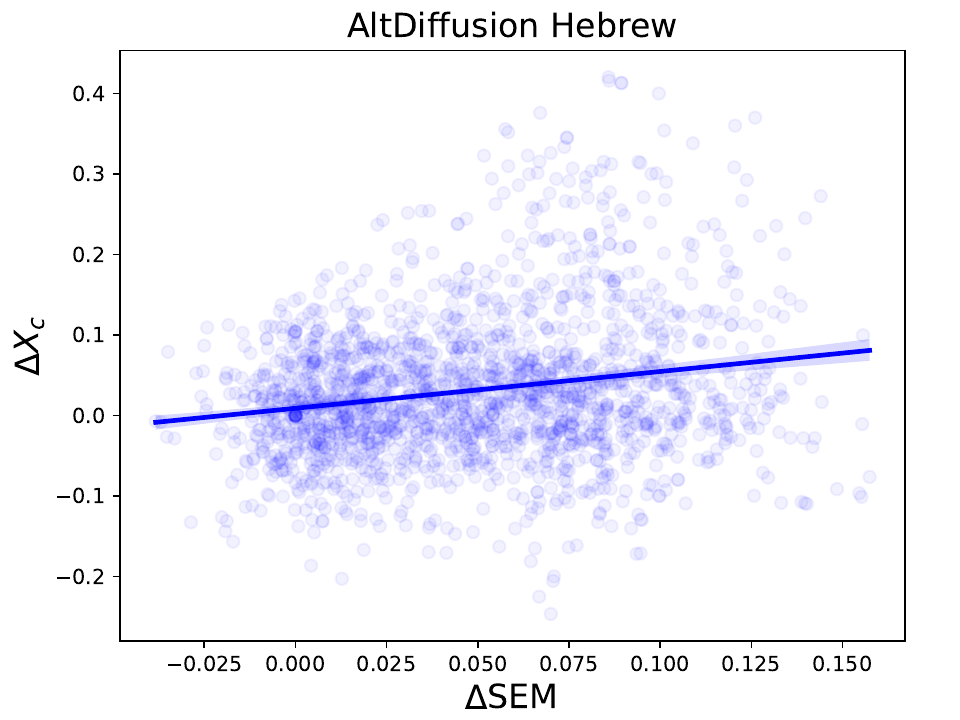} 
\caption{Scatterplots for the pseudocorrection experiments. Transparent circles are used to make distribution mass more visible.}
\label{fig:scatter_rand}
\end{figure*}

\begin{table*}[t]\centering
\small

\begin{tabular}{llll|rrrrrr}\toprule
\multirow{2}{*}{Concept}&\multirow{2}{*}{Original} &\multirow{2}{*}{Corrected} &\multirow{2}{*}{Type} &\multirow{2}{*}{$\Delta\mathrm{SEM}$} & \multicolumn{4}{c}{$\Delta X_c$ (CCCL Improvement) for model}\\
&&&&&SD 1.4&SD 2 &SD 2.1 &AD \\\cmidrule{1-9}
\multicolumn{9}{c}{\textit{All Japanese-language error candidates:}}\\\cmidrule{1-9}
duck &\inlinejp{鴨} &\inlinejp{アヒル} & C &-0.092 &0.021 &0.008 &-0.012 &-0.055 \\
thigh &\inlinejp{腿} &\inlinejp{ふともも} & C &-0.091 &0.048 &0.007 &-0.043 &-0.124 \\
cop &\inlinejp{警官} &\inlinejp{お巡りさん} & F & -0.053 &-0.160 &-0.029 &-0.055 &-0.140 \\
field &\inlinejp{分野} &\inlinejp{田んぼ} & A &-0.036 &0.015 &-0.151 &-0.075 &-0.058 \\
butterfly &\inlinejp{蝶} &\inlinejp{蝶々} & C &-0.022 &-0.004 &0.025 &0.009 &-0.020 \\
girlfriend &\inlinejp{ガールフレンド} &\inlinejp{彼女} & C & -0.013 &0.044 &0.166 &0.196 &-0.030 \\
stingray &\inlinejp{アカエイ} &\inlinejp{エイ} & C &-0.008 &-0.058 &0.044 &-0.006 &-0.071 \\
cigarette &\inlinejp{煙草} &\inlinejp{たばこ} & C &-0.007 &0.054 &0.043 &-0.034 &0.078 \\
tail &\inlinejp{尾} &\inlinejp{尻尾} & C &-0.003 &0.004 &0.077 &0.056 &0.040 \\
woman &\inlinejp{女性} &\inlinejp{女} & C &-0.001 &0.108 &-0.022 &-0.014 &-0.046 \\
forest &\inlinejp{森林} &\inlinejp{森} & C &-0.000 &0.226 &0.081 &0.032 &0.051 \\
teenager &\inlinejp{ティーンエイジャー} &\inlinejp{少年} & C, T & 0.002 &0.169 &0.076 &0.115 &0.023 \\
flame &\inlinejp{火炎} &\inlinejp{炎} & C &0.003 &-0.062 &-0.070 &0.009 &0.031 \\
father &\inlinejp{父} &\inlinejp{父親} & F &0.010 &-0.009 &-0.010 &0.014 &0.003 \\
watch &\inlinejp{時計} &\inlinejp{腕時計} & IS & 0.011 &0.487 &0.080 &0.062 &0.006 \\
teacher &\inlinejp{先生} &\inlinejp{教師} & IS & 0.015 &0.006 &-0.051 &-0.070 &0.016 \\
kid &\inlinejp{キッド} &\inlinejp{子ども} & C, T & 0.017 &0.098 &0.070 &0.065 &0.068 \\
doctor &\inlinejp{先生} &\inlinejp{医者} & IS & 0.017 &-0.006 &0.031 &0.018 &0.050 \\
ground &\inlinejp{接地} &\inlinejp{地面} & OS & 0.022 &-0.008 &0.097 &0.084 &0.086 \\
bike &\inlinejp{バイク} &\inlinejp{自転車} & OS, T & 0.023 &0.195 &0.021 &-0.018 &0.020 \\
detail &\inlinejp{ディテール} &\inlinejp{詳細} & C, T & 0.024 &0.002 &0.036 &0.043 &-0.031 \\
milk &\inlinejp{乳} &\inlinejp{牛乳} & OS & 0.033 &0.141 &0.026 &-0.002 &0.215 \\
cafeteria &\inlinejp{カフェテリア} &\inlinejp{食堂} & C, T &0.044 &-0.192 &-0.043 &-0.034 &0.064 \\
rock &\inlinejp{ロック} &\inlinejp{岩} & IS, T & 0.067 &0.048 &-0.029 &-0.033 &0.104 \\
\cmidrule{1-9}
\multicolumn{9}{c}{\textit{All Chinese-language error candidates:}}\\\cmidrule{1-9}
men &\inlinezh{男人} &\inlinezh{很多人} & A &-0.032 &0.001 &-0.180 &-0.182 &-0.411 \\
stingray &\inlinezh{黄貂鱼} &\inlinezh{鳐鱼} & C &-0.030 &0.082 &0.206 &0.213 &-0.099 \\
field &\inlinezh{领域} &\inlinezh{田野} & A &-0.017 &-0.012 &-0.136 &-0.184 &0.083 \\
boat &\inlinezh{船} &\inlinezh{小船} & F &-0.001 &-0.110 &0.009 &0.008 &0.017 \\
sister &\inlinezh{姐姐} &\inlinezh{姐妹} & F &-0.001 &0.033 &0.014 &0.026 &-0.014 \\
wife &\inlinezh{老婆} &\inlinezh{妻子} & C &0.003 &-0.021 &0.124 &0.177 &-0.021 \\
bottle &\inlinezh{瓶} &\inlinezh{瓶子} & C &0.004 &-0.062 &-0.021 &0.032 &0.075 \\
church &\inlinezh{教会} &\inlinezh{教堂} & A &0.005 &-0.068 &0.076 &0.078 &-0.018 \\
father &\inlinezh{爸爸} &\inlinezh{父亲} & C &0.009 &0.027 &-0.028 &-0.059 &0.145 \\
mouth &\inlinezh{口} &\inlinezh{嘴} & C &0.011 &-0.054 &0.023 &0.010 &0.037 \\
bell &\inlinezh{钟} &\inlinezh{铃} & A &0.013 &-0.013 &0.071 &0.081 &-0.001 \\
cafeteria &\inlinezh{自助餐厅} &\inlinezh{食堂} & A &0.017 &-0.102 &-0.047 &-0.054 &0.071 \\
orange &\inlinezh{橙色} &\inlinezh{橙子} & OS & 0.019 &0.002 &-0.099 &-0.104 &0.067 \\
belt &\inlinezh{带} &\inlinezh{皮带} & IS &0.029 &0.025 &0.045 &0.034 &0.040 \\
suit &\inlinezh{适合} &\inlinezh{西装} & OS &0.033 &-0.003 &-0.062 &-0.052 &0.329 \\
hallway &\inlinezh{门厅} &\inlinezh{走廊} & A &0.045 &0.166 &0.011 &0.015 &0.105 \\
table &\inlinezh{表} &\inlinezh{桌子} & OS &0.064 &-0.068 &0.098 &0.043 &0.206 \\
\cmidrule{1-9}
\multicolumn{9}{c}{\textit{All Spanish-language error candidates:}}\\\cmidrule{1-9}
ticket &boleto &billete & C &-0.034 &0.169 &0.036 &0.069 &0.011 \\
room &habitación &cuarto & C &-0.005 &-0.184 &-0.166 &-0.094 &-0.083 \\
bird &pájaro &ave & C &-0.001 &-0.437 &-0.373 &-0.433 &-0.020 \\
flame &llama &flama & T, C &0.004 &-0.040 &-0.134 &-0.164 &0.044 \\
ship &navío &barco & C &0.005 &0.002 &0.132 &0.149 &-0.083 \\
hill &cerro &colina & C &0.019 &-0.023 &-0.005 &-0.116 &0.078 \\
kid &cabrito &joven & C, F &0.022 &0.027 &0.077 &0.065 &0.100 \\
tent &tienda &tienda de acampar & A, IS & 0.072 &-0.005 &0.013 &-0.013 &0.353 \\
sandwich &emparedado &sándwich & C &0.098 &0.254 &0.519 &0.534 &0.339 \\
\bottomrule
\end{tabular}
\caption{All identified concept translation error candidates in the original CoCo-CroLa and their corresponding corrections in Japanese, Chinese, and Spanish. Each section is sorted in ascending order of $\Delta\mathrm{SEM}$. Error types are defined in \autoref{sec:errtype}}\label{tab: mistranslation-examples-all}
\end{table*}

\end{document}